\documentclass{article}

\usepackage[preprint]{corl_2026} 

\usepackage{booktabs}
\usepackage[table]{xcolor}
\usepackage{multirow}
\usepackage{booktabs}
\usepackage{array}
\usepackage{subcaption}
\usepackage{caption}
\usepackage{booktabs}
\usepackage{amsmath}
\usepackage{makecell}
\usepackage{amssymb}
\usepackage{wrapfig}
\usepackage{pifont}
\usepackage{cleveref}

\newcommand{\xmark}{\ding{55}}
\usepackage{xcolor}
\usepackage{graphicx}
\definecolor{lowgray}{gray}{0.55}

\usepackage{lipsum}
\usepackage{graphicx}
\usepackage{xcolor}
\newcommand\blfootnote[1]{
        \begingroup
        \renewcommand\thefootnote{}\footnote{#1}
        \addtocounter{footnote}{-1}
        \endgroup
    }
\definecolor{g_sum}{HTML}{3479E5}

\title{Geometric Action Model for Robot Policy Learning}

\author{
  Jisang Han$^{*\,1}$ \quad Seonghu Jeon$^{*\,1}$ \quad Jaewoo Jung$^{1,2}$ \quad René Zurbrügg$^{2,3}$\\
  \textbf{Honggyu An$^{1}$
  \quad Tifanny Portela$^{2,3}$
  \quad Marco Hutter$^{2}$}\\
  \textbf{Marc Pollefeys$^{2}$ \quad Seungryong Kim$^{\dagger\, 1}$ \quad Sunghwan Hong$^{\dagger \, 2,3}$} \\[8pt]
  $^{1}$KAIST AI \quad $^{2}$ETH Zurich \quad $^{3}$ETH AI Center\\[2pt]
  {\tt \href{https://cvlab-kaist.github.io/Geometric-Action-Model/}{\textcolor{g_sum}{https://cvlab-kaist.github.io/Geometric-Action-Model}}}
  \vspace{-25pt}
}

\begin{document}
\maketitle
\blfootnote{$^*$Equal contribution.}
\blfootnote{$^\dagger$Corresponding author.}



\vspace{-20px}
\begin{figure}[h]
    \centering
    \includegraphics[width=1\linewidth]{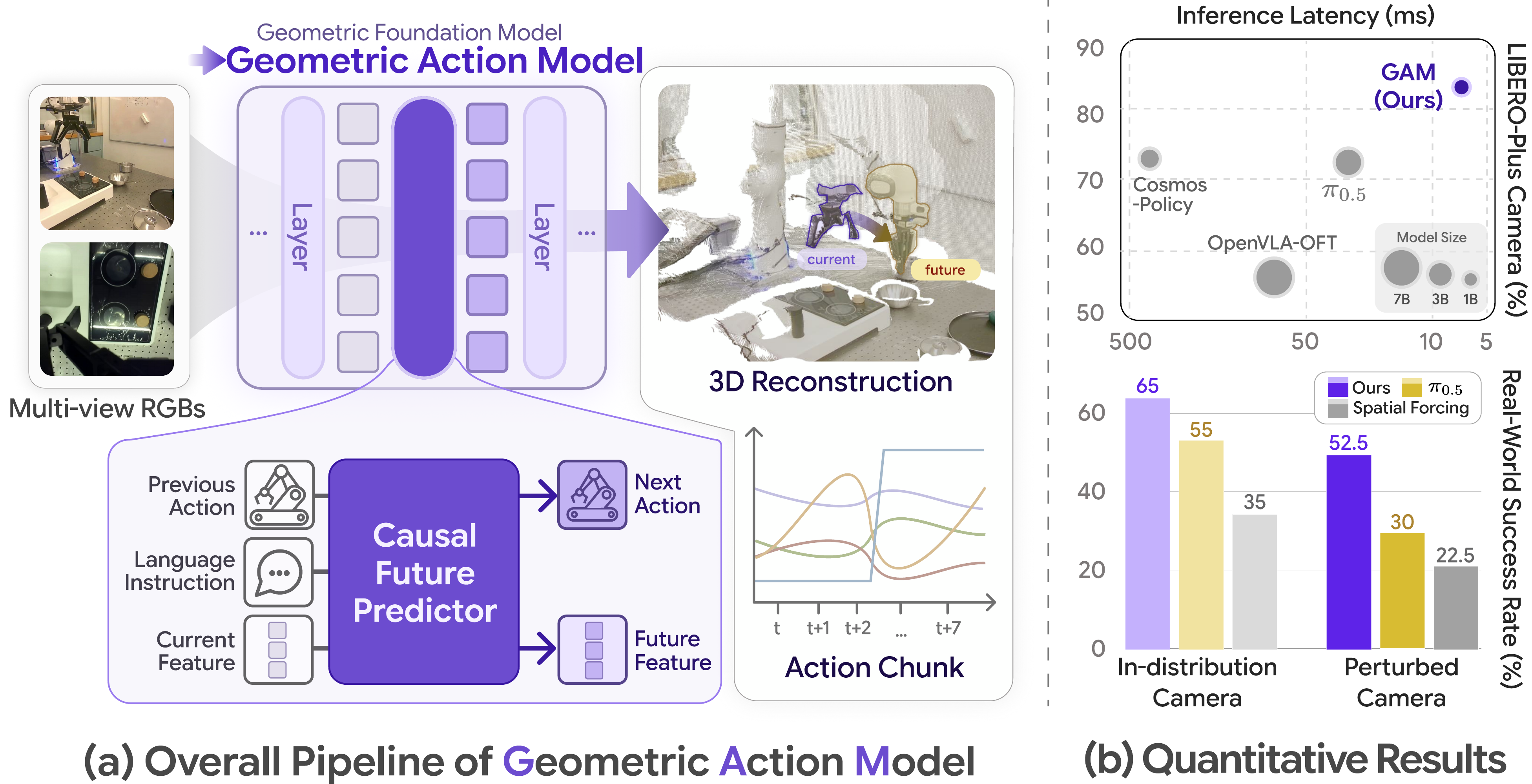}
    \vspace{-10px}
    \caption{\textbf{GAM repurposes geometric foundation models into fast and robust robot policies.} (a) GAM jointly predicts future 3D geometry and action chunks within a shared geometric backbone. (b) By leveraging explicit 3D geometric priors, GAM improves robustness and real-world performance while reducing latency and model size compared to existing baselines.}
    \vspace{-3pt}
    \label{fig:intuition}
\end{figure}

\begin{abstract}
Generalist robot policies must follow user instructions while reasoning about how objects, cameras, and robot actions interact in the 3D physical world. Recent vision-language-action models (VLAs) and video world-action models (WAMs) inherit strong semantic or temporal priors from large-scale foundation models, but they still operate primarily on 2D image frames or 2D-derived latent spaces, leaving implicit the 3D geometry required for contact-rich manipulation. We propose the \textbf{Geometric Action Model (GAM)}, a language-conditioned manipulation policy that directly repurposes a pretrained geometric foundation model (GFM) as a shared substrate for perception, temporal prediction, and action decoding. GAM splits the GFM at an intermediate layer: the shallow layers serve as an observation encoder, and a causal future predictor inserted at the split layer forecasts future latent tokens conditioned on language, proprioception, and action history. The predicted future tokens are then routed through the remaining GFM blocks for feature propagation and decoding, allowing a single backbone to produce both future geometry and actions. This design equips the GFM with language-conditioned temporal world modeling through minimal architectural modification while preserving its rich geometric priors. Across a broad suite of simulation and real-robot manipulation benchmarks, GAM is more accurate, more robust, faster, and lighter than current foundation-model-scale baselines.
\end{abstract}
\section{Introduction}
\label{sec:intro}
A long-standing goal in robotics is to build generalist manipulation policies that can follow natural-language instructions and manipulate arbitrary objects across diverse scenes~\citep{kim2024openvla, openvla_oft, kim2026cosmos, black2024pi_0}. To achieve this, a general manipulation model must not only recognize objects and parse instructions, but also reason about how the physical world will evolve under its own actions. This requires a unified understanding of language, visual appearance, scene geometry, robot state, and physical dynamics. 

Recent progress has therefore increasingly relied on large-scale foundation models as pretrained substrates for robot policies. Vision-language-action models (VLAs) build on vision-language models whose representations are aligned with natural language, and learn to map visual and linguistic tokens to robot actions~\cite{zitkovich2023rt,kim2024openvla,openvla_oft,black2024pi_0,pi05,pi0_fast}. Video world-action models (WAMs) instead leverage pretrained video generation models, using their world prediction priors to jointly model future frames and actions~\cite{kim2026cosmos,pai2025mimic}. While these approaches have shown impressive language-conditioned manipulation ability, they are fundamentally in 2D: 3D cues such as depth, scale, and occlusion are left implicit in monocular cues that the action decoder must disentangle on its own, leading to limited generalization across environment changes, especially in robot initial state and camera viewpoint~\cite{fei2025libero, wilcox2025adapt3r}. 

To overcome this limitation, recent work incorporates 3D geometric information into robot policies. One line learns policies directly on explicit 3D observations such as raw point clouds~\citep{ze20243d,huang2026pointworld}, demonstrating the value of geometry for generalization but typically requiring task-specific encoders trained from scratch. With the emergence of Geometric Foundation Models (GFMs)~\cite{lin2025depth,wang2025vggt,wang2026vggt}, some works transfer pretrained geometric priors into VLA policies, either distilling selected GFM features into the VLA backbone through representation alignment~\citep{yu2024representation,li2025spatial,sun2026rocket} or attaching a lightweight action head on top of a GFM's final features~\citep{qian2025gp3,ge2025vggt}. These improve spatial awareness, but use the GFM only as a static feature extractor: its multi-layer geometric structure is never repurposed as the policy's own temporal and action-generating substrate.  

In this work, we propose \textbf{Geometric Action Model (GAM)}, which directly repurposes a GFM as a manipulation policy by using it as a shared medium for perception, future-state prediction, and action decoding. We show that by jointly predicting future action and geometry, geometric world dynamics can be inherently incorporated into robot policies.

Specifically, we split the pretrained GFM at an intermediate layer: the shallow layers serve as an observation encoder, while the remaining layers serve as a decoder block. Given the current visual observation, the observation encoder extracts spatially meaningful scene representations. To model how the world evolves over time, we insert a causal transformer at the intermediate layer that predicts future feature representations. This predictor is conditioned on task language, proprioception, and action history by introducing them as additional tokens at each timestep. The predicted future-state tokens are then processed by the remaining GFM decoder together with an action token, allowing the backbone to produce both future geometry and robot actions. An intuitive comparison between existing paradigms and our proposed framework is illustrated in Figure~\ref{fig:intuition}.

Across diverse simulation~\citep{liu2023libero,fei2025libero,nasiriany2024robocasa} and real-world benchmarks, GAM matches or exceeds the success rate of current foundation-model-scale baselines such as VLAs and WAMs while using substantially fewer trainable parameters and substantially faster inference \textbf{(55$\times$ faster)}, and shows improved generalization to unseen scenarios. GAM especially achieves outstanding performance in camera perturbation settings \textbf{($\uparrow$9.7\%p)}, which requires geometric understanding priors. 

Our contributions are as follows:
\vspace{-2.5pt}
\begin{itemize}
    \item We introduce \textbf{Geometric Action Model (GAM)}, a manipulation policy that combines temporal world modeling, latent feature-space prediction, and a geometric foundation-model substrate in a single shared-backbone architecture.
    \item We show that action and geometry can be predicted in a shared token space: a single autoregressive sequence and a single backbone forward pass produce both action tokens and future-scene tokens, decoded by a lightweight action regression head and a depth head.
    \item We demonstrate that across diverse simulation and real-world manipulation benchmarks, GAM is simultaneously more accurate, more robust, faster, and lighter than current foundation-model-scale alternatives.
\end{itemize}
\section{Related Work}
\label{sec:related}
\begin{figure}[t]
    \centering
    \includegraphics[width=1\linewidth]{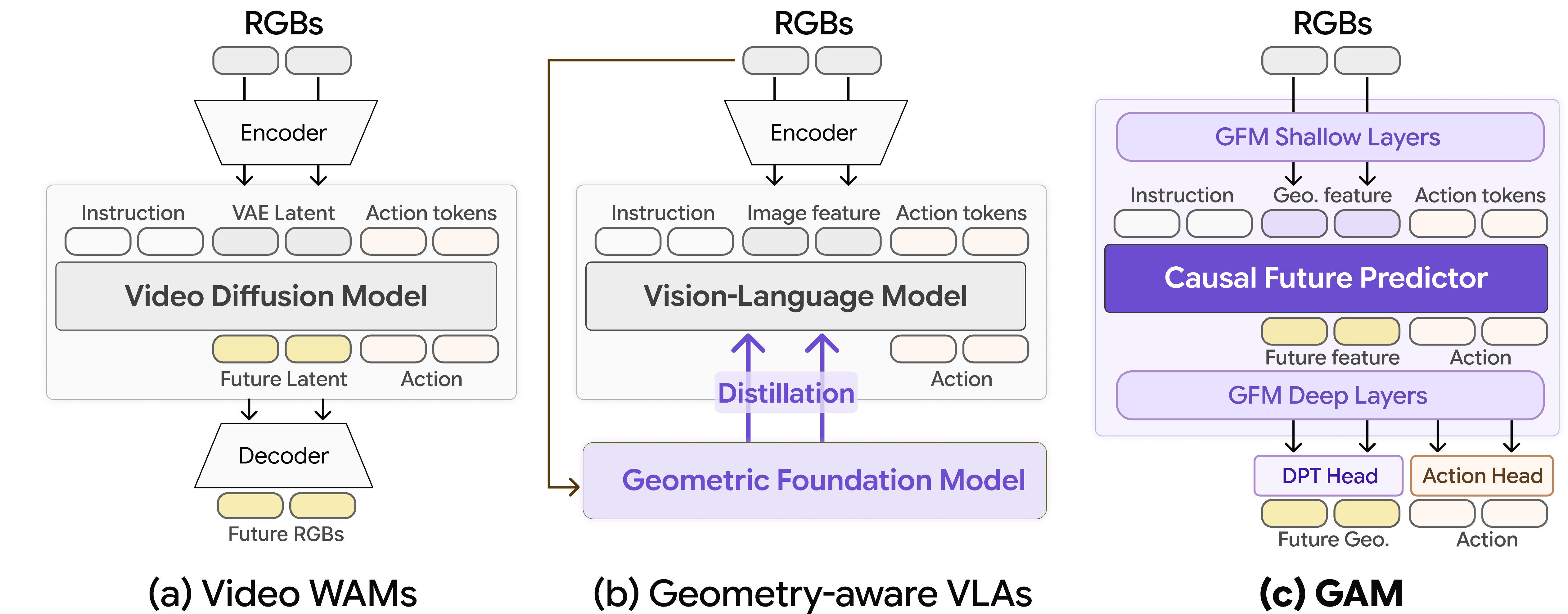}
    \caption{\textbf{(a) Video WAMs}~\citep{kim2026cosmos,pai2025mimic, ye2026world} operate in 2D pixel space, predicting future latents and actions via video diffusion. \textbf{(b) Geometry-aware VLAs}~\citep{li2025spatial, sun2026rocket} predict actions using a VLA with passive feature distillation from an external GFM. \textbf{(c) GAM (ours)} unifies perception, geometry prediction, and action decoding by inserting a geometric world model inside a single GFM.}

    \vspace{-10pt}
    \label{fig:intuition}
\end{figure}
\noindent\textbf{Vision-Language-Action Models.}  
Vision-language-action models (VLAs) adapt a pretrained vision-language model (VLM) into a robot policy. Early works~\citep{zitkovich2023rt, kim2024openvla} autoregressively decode discrete action tokens from a finetuned VLM. This paradigm is extended through parallel decoding~\citep{openvla_oft}, flow-matching action experts~\citep{black2024pi_0, pi05, shi2025hi}, diffusion-based action heads~\citep{liu2025rdt}, frequency-space action tokenizations~\citep{pi0_fast}, and compact open-source VLMs~\citep{nora,bai1others}.
This line of work extends large generalist foundation models for humanoid and embodied control~\citep{bjorck2025gr00t, team2025gemini, cheang2025gr, yang2025magma}, to spatial-representation variants~\citep{qu2025spatialvla}, and to refinements in training and post-training ~\citep{univla, riptvla, worldvla}.
While these VLAs leverage strong open-vocabulary recognition to produce policies, they rely solely on 2D image priors and therefore lack 3D understanding. To address this, recent works~\citep{li2025spatial, sun2026rocket, abouzeid2025geoaware} attempt to align intermediate VLA features with geometric foundation model. However, they do not fully exploit the geometric understanding of geometric foundation model backbone.
 
\noindent\textbf{World Action Models for Robot Manipulation.}
World action models are trained to predict future states in order to learn a policy. One branch builds on large pretrained video generation models~\citep{agarwal2025cosmos}, fine-tuning them to jointly predict future frames and actions~\citep{kim2026cosmos, pai2025mimic, ye2026world}, arguing that the benefit of such co-training stems primarily from training-time supervision rather than test-time future imagination~\citep{ye2026world, pi0_fast}. A second branch keeps the visual backbone frozen and trains a separate temporal predictor in its feature space for planning via model-predictive control~\citep{zhou2024dino, assran2025v}, with related work using implicit future-latent alignment as an auxiliary training signal~\citep{zheng2025flare}. However, these video backbones encode only 2D image-space priors, which do not explicitly resolve depth, scale, or occlusion. GAM instead predicts in the latent space of a geometric foundation model that encodes rich 3D priors, while inheriting from both branches the paradigm of using future prediction as a training signal.

\noindent\textbf{Geometric Foundation Models for Manipulation.} Geometric foundation models (GFMs)~\cite{wang2025vggt, lin2025depth, wang2025pi, keetha2025mapanything} infer dense 3D structures from multi-view images and have recently served as perceptual substrates for robot policies. Early approaches integrate GFMs with VLAs as frozen feature extractors via representation alignment~\citep{li2025spatial, sun2026rocket}, direct encoder replacement~\citep{abouzeid2025geoaware, qian2025gp3}, or point-cloud fusion~\citep{sun2025geovla}. Moving beyond static extraction, recent concurrent works adopt GFMs for predictive control: \citet{song2026robotic} utilizes the GFM to jointly predict actions and \emph{current}-frame 3D properties, while \citet{xu2026action} employs the GFM with a diffusion policy to co-denoise future action chunks and 3D latents. GAM departs from these paradigms in two ways: (1) action and future-scene predictions are jointly modeled within a single autoregressive token sequence rather than separated heads or diffusion processes, and (2) the GFM's deep blocks are explicitly repurposed to decode predicted \emph{future} tokens rather than merely processing observed ones.

\section{Preliminaries: Geometric Foundation Models}
\label{sec:prelim}

A geometric foundation model (GFM) such as VGGT~\citep{wang2025vggt} or DA3~\citep{lin2025depth} is a feed-forward transformer that maps one or more RGB images to dense 3D geometry. Given a sequence of $V$ views $\mathcal{I} = \{I_v\}_{v=1}^{V}$ with $I_v \in \mathbb{R}^{3 \times h \times w}$, a GFM produces per-pixel depth $D_v \in \mathbb{R}^{h \times w}$ or 3D point maps $P_v \in \mathbb{R}^{3 \times h \times w}$ in a shared world frame, and per-view camera intrinsics and extrinsics $(K_v, \xi_v) \in \mathbb{R}^{3\times 3} \times SE(3)$ via auxiliary heads.
Specifically, each view $I_v$ is partitioned into $P$ non-overlapping patches of size $p \times p$ and projected by a patch embedding into a per-view token sequence:
\begin{equation}
    \mathbf{z}_v^{(0)} = \big[\mathbf{c}_v,\; \mathbf{x}_v^1, \ldots, \mathbf{x}_v^{P}\big] \in \mathbb{R}^{(1+P) \times d},
\end{equation}
where $\mathbf{c}_v$ is a per-view camera token, $\{\mathbf{x}_v^j\}_{j=1}^{P}$ are patch tokens, and $d$ is the hidden dimension. The full input sequence concatenated across views is $\mathbf{Z}^{(0)} = [\mathbf{z}_1^{(0)}, \ldots, \mathbf{z}_V^{(0)}] \in \mathbb{R}^{V(1+P) \times d}$.

These tokens are processed by a stack of $M$ transformer blocks $\{{f^{(m)}}\} _{m=1}^M$ employing one of two attention modes. \emph{Frame-wise attention} $f_{\text{frame}}^{(m)}$ operates within each view tokens independently, attending over the $(1+P)$ tokens of a single image. \emph{Global attention} $f_{\text{global}}^{(m)}$ operates jointly over all $V (1+P)$ tokens, fusing information across viewpoints. The hidden state at the $m$-th transformer block evolves as follows:
\begin{equation}
    \mathbf{Z}^{(m)} = f^{(m)}\big(\mathbf{Z}^{(m-1)}\big), \quad f^{(m)} \in \{f_{\text{frame}}^{(m)},\, f_{\text{global}}^{(m)}\}.
\end{equation}
After the transformer forward pass, dense geometry is decoded from multiple intermediate hidden states $\mathbf{Z}^{(m^*)}$, where $m^*$ is one of the selected layers in $\mathcal{S}=\{m_1,m_2,m_3,m_4\}$. The extracted multi-layer hidden states are then fed into the DPT~\cite{ranftl2021vision} head to estimate per-pixel geometry.

\section{GAM: Geometric Action Model}
\label{sec:method}

\begin{figure}
    \centering
    \includegraphics[width=1\linewidth]{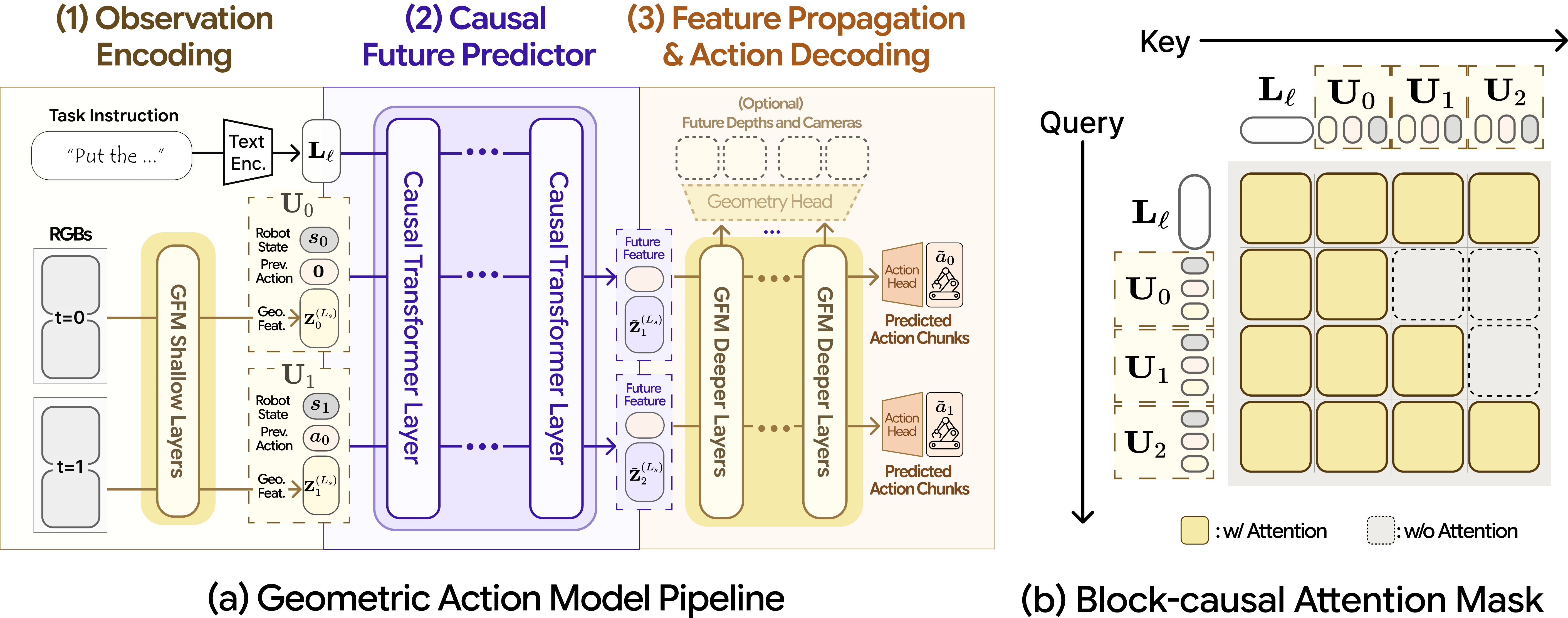}
  \caption{\textbf{Main architecture of GAM.} }
  \vspace{-15pt}

    \label{fig:arch}
\end{figure}

\noindent\textbf{Problem Formulation.} We consider language-conditioned robot manipulation. At each timestep $t$, the robot receives a multi-view RGB observation $o_t = \{I_{v,t}\}_{v=1}^{V}$ from $V$ fixed cameras, a proprioceptive state $s_t \in \mathbb{R}^{d_s}$ describing the robot's joint configuration and end-effector pose, and a natural-language task instruction $\ell$ that is held constant throughout an episode. The policy $\pi_\theta$ must produce an action chunk $\hat{a}_t \in \mathbb{R}^{C \times d_a}$ of length $C$, encoding the next $C$ delta-pose or joint commands to be executed open-loop before the next observation is acquired. We learn a policy:
\begin{equation}
    \pi_\theta\colon\;\big(\{o_{t-H+1}, \ldots, o_t\},\, \{s_{t-H+1}, \ldots, s_t\},\, \{a_{t-H}, \ldots, a_{t-1}\},\, \ell\big)\;\mapsto\;\hat a_t
\end{equation}
from a dataset of $N$ expert demonstrations $\mathcal{D} = \{(\tau_i, \ell_i)\}_{i=1}^N$, where each trajectory $\tau_i = (o_t, s_t, a_t)_{t=1}^{T_i}$ pairs a sequence of observations, states, and executed action chunks with a fixed instruction $\ell_i$. The policy conditions on a context window of $H$ recent timesteps.

\noindent\textbf{Overview.} In the following sections, we explain how we transform a pretrained GFM into a
language-conditioned world-action model. Our key idea is to split the GFM into two parts and insert a causal temporal predictor between them. This design lets GAM formulate future prediction directly inside the GFM latent space, enabling all predictive computation to be performed in the GFM's geometric representation space.

Concretely, our framework operates in three sequential stages inside the GFM. First, the \textbf{observation encoder} (§\ref{sec:method:encoding}) repurposes the shallow layers of the GFM to extract latent geometric features from multi-view observations. Next, the \textbf{causal future predictor} (§\ref{sec:method:predictor}) operates at the split layer, where it combines these geometric features with language, proprioception, and action history to
predict future latent tokens. Finally, during \textbf{feature propagation and decoding} (§\ref{sec:method:decoding}), the predicted future tokens are routed through the remaining deep GFM blocks to simultaneously decode future geometry and the final action chunk $\hat{a}_t$. Figure~\ref{fig:arch} (a) shows the overall architecture of our model.

\subsection{Observation Encoder}
\label{sec:method:encoding}
We first reuse the shallow layers of the pretrained GFM as the observation encoder. Let $L_s$ denote the split layer where the causal future predictor is inserted. The original GFM transformer stack is then decomposed into an encoder and a decoder:
\begin{equation}
    E_{\leq L_s}
    =
    f^{(L_s)}
    \circ
    \cdots
    \circ
    f^{(1)},
    \qquad
    D_{> L_s}
    =
    f^{(M)}
    \circ
    \cdots
    \circ
    f^{(L_s+1)} .
\end{equation}
Here, the choice of $L_s$ is important because $L_s$ must be deep enough to extract sufficiently rich visual features from the raw observations, yet shallower than the earliest layer used in the DPT head $L_s < m_1$, so that predicted future states can be decoded into future geometries by the DPT heads.

After defining this split layer $L_s$, for each timestep $t'$ in the context window, we tokenize the multi-view RGB observation $o_{t'}=\{I_{v,t'}\}_{v=1}^{V}$ using the original GFM patch embedding. This produces the initial multi-view token sequence:
\begin{equation}
    \mathbf{Z}_{t'}^{(0)}
    =
    \big[
        \mathbf{z}_{1,t'}^{(0)},\ldots,
        \mathbf{z}_{V,t'}^{(0)}
    \big]
    \in
    \mathbb{R}^{V(1+P)\times d},
\end{equation}
where each view contributes one camera token and $P$ patch tokens. The observation encoder maps these tokens to the split-layer representation $\mathbf{Z}_{t'}^{(L_s)}$. By applying this encoding independently to each timestep in the context window, the output of this stage is a sequence of per-timestep geometric latent states $\{\mathbf{Z}_{t-H+1}^{(L_s)},\ldots,\mathbf{Z}_{t}^{(L_s)}\}$.

\subsection{Causal Future Predictor}
\label{sec:method:predictor}

After the observation encoder, GAM performs temporal prediction directly at the split layer $L_s$, forecasting the next latent geometric state from current and past observations while conditioning on the task instruction, proprioception, and action history. To this end, we insert a causal future predictor $g_\phi$ between the shallow encoder $E_{\leq L_s}$ and the deep decoder $D_{>L_s}$.
For each timestep $t'$ in the context window, the encoder provides latent tokens $\mathbf{Z}_{t'}^{(L_s)}$, and we embed the proprioceptive state $s_{t'}$ and previous action $a_{t'-1}$ as tokens:
\begin{equation}
    \mathbf{p}_{t'} = \psi_s(s_{t'}), 
    \qquad
    \mathbf{q}_{t'} = \psi_a(a_{t'-1}),
\end{equation}
with $\psi_s, \psi_a$ lightweight projection layers, and the instruction $\ell$ into language tokens $\mathbf{L}_{\ell}$ with a pretrained text encoder. We then form a per-timestep token block by concatenating the encoded GFM tokens with the proprioception and action-history tokens $\mathbf{U}_{t'}=[\mathbf{p}_{t'};\mathbf{q}_{t'};\mathbf{Z}_{t'}^{(L_s)}]$. The full input to the causal future predictor is $\mathbf{X} = [\mathbf{L}_\ell; \mathbf{U}_{t'-H+1}; \ldots; \mathbf{U}_{t'}]$.

The combined sequence $\mathbf{X}$ is then processed through block-causal self-attention~\citep{wang2025pi}, ensuring the model incorporates past and present contexts without future leakage, as illustrated in Figure~\ref{fig:arch} (b). At the final layer of the predictor $g_\phi$, we read off the predictions from their respective sequence slots. Specifically, the hidden states corresponding to the geometry slots forecast the latent geometric tokens of the future frame, denoted as $\tilde{\mathbf{Z}}_{t'+1}^{(L_s)}$. Concurrently, the hidden state of the designated previous-action slot is projected to produce a predicted next action token $\tilde{\mathbf{a}}_{t'} \in \mathbb{R}^{d}$, in direct analogy to next-token prediction in a causal language model. By jointly forecasting action and geometric latents in this layer, we ensure that action tightly interacts with spatial representations.

This design of introducing a causal transformer predictor $g_\phi$ allows the pretrained GFM to acquire language-conditioned temporal world modeling with \textit{minimal} architectural modification. Only the inserted $g_\phi$ needs to learn how to fuse language, proprioception, and action history with GFM latent features. The resulting predictions, $\widetilde{\mathbf{Z}}_{t'+1}^{(L_s)}$ and $\tilde{\mathbf{a}}_{t'}$, are then passed to the remaining GFM blocks for joint geometry and action decoding.

\subsection{Feature Propagation and Action Decoding}
\label{sec:method:decoding}

Following the causal future predictor, the single action token $\tilde{\mathbf{a}}_{t'}$ is replicated $V$ times to form a set of per-view action tokens $\{\tilde{\mathbf{a}}_{v,t'}\}_{v=1}^{V}$, where $\tilde{\mathbf{a}}_{v,t'} = \tilde{\mathbf{a}}_{t'}$. Concatenated with the geometry tokens, they are fed through the remaining GFM blocks $D_{> L_s}$. We perform this \emph{feature propagation} by appending each view's corresponding action token $\tilde{\mathbf{a}}_{v,t'}$ directly to its geometry token sequence for each timestep:
\begin{equation}
    \tilde{\mathbf{Z}}_{t'+1}^{(M)} = \big(f^{(M)} \circ \cdots \circ f^{(L_s+1)}\big)\!\Big(\Big[\big[\tilde{\mathbf{Z}}_{1,t'+1}^{(L_s)};\, \tilde{\mathbf{a}}_{1,t'}\big], \ldots, \big[\tilde{\mathbf{Z}}_{V,t'+1}^{(L_s)};\, \tilde{\mathbf{a}}_{V,t'}\big]\Big]\Big).
\end{equation}
To prevent future leakage, we extend the predictor's causal mask strategy to the GFM's remaining global attention layers ($f_{\text{global}}^{(m)}$).

Finally, the propagated features are decoded by two heads. The lightweight action head $h_{\text{act}}$ aggregates action tokens over the context window to regress the executable action chunk $\hat a_{t'}$, while the original GFM depth head $h_{\text{depth}}$ decodes geometry tokens into action-aligned future depth maps. The GFM's deep blocks, originally pretrained to decode shallow features into 3D geometry, are thus repurposed here as the decoder of the world model's predicted future.

\subsection{Training and Inference}
\label{sec:method:training}
The policy is trained end-to-end by minimizing a multi-task objective over action execution, world modeling, and geometric decoding:
\begin{equation}
    \mathcal{L}_{\text{total}} = \lambda_{\text{act}} \mathcal{L}_{\text{act}} + \lambda_{\text{feat}} \mathcal{L}_{\text{feat}} + \lambda_{\text{depth}} \mathcal{L}_{\text{depth}},
\end{equation}
where the $\lambda$ factors balance each term and $\mathcal{H} = \{t-H+1, \ldots, t\}$ is the context window. The \emph{action} loss $\mathcal{L}_{\text{act}}$ is an $\ell_1$ regression between the decoded action chunk $\hat a_{t'}$ and the expert action $a_{t'}$ over all $t' \in \mathcal{H}$. The \emph{future-feature} loss $\mathcal{L}_{\text{feat}}$ anchors the predictor $g_\phi$ to temporal geometric transitions by aligning predicted future tokens $\tilde{\mathbf{Z}}_{t'+1}^{(L_s)}$ with the actual next frame $\mathbf{Z}_{t'+1}^{(L_s)}$ extracted from frozen GFM:
\begin{equation}
    \mathcal{L}_{\text{feat}} = \sum_{t' \in \mathcal{H}} \left\| \tilde{\mathbf{Z}}_{t'+1}^{(L_s)} - \mathbf{Z}_{t'+1}^{(L_s)} \right\|_1.
\end{equation}
The \emph{future-depth} loss $\mathcal{L}_{\text{depth}}$ grounds the predicted future in valid 3D structure by supervising the decoded depth $\tilde{D}_{t'+1}=h_{\text{depth}}(\tilde{\mathbf{Z}}_{t'+1}^{(m^*)})$ using depth head $h_{\text{depth}}$ against ground-truth future depth $D_{t'+1}$, adopting the scale-invariant and gradient-matching penalties of the GFM~\citep{lin2025depth, wang2026vggt}. 

At inference, we maintain the historical context online with key-value caching, so each step processes only the new observation $o_t$ and previous action $a_{t-1}$ in a single feed-forward pass. 

\section{Experiments}
\label{sec:exp}
\subsection{Implementation Details}

We use DA3-Giant~\citep{lin2025depth} fine-tuned on Track4World~\citep{lu2026track4world} as the backbone. We insert a 12-layer causal predictor with width $d_g=1024$ at layer $L_s=12$, where alternating attention begins. For the task instruction, we extract language tokens using a frozen T5 encoder~\citep{raffel2020exploring}. The policy uses a context horizon of $H=4$ for pre-training and $H=1$ for post-training and predicts $C=8$ step action chunks in a $d_a=7$ end-effector action space from $d_s=7$ proprioceptive states. We pretrain GAM on 784K single-arm robot trajectories from RoboCasa365~\citep{nasiriany2026robocasa365}, MimicGen~\citep{mandlekar2023mimicgen}, and OpenX-Embodiment~\citep{collaboration2023open}, then post-train it on each benchmark. We optimize with AdamW using a constant learning rate, freeze layers before $L_s$ and the depth head, and supervise depth with simulator ground truth. We set $\lambda_{\text{act}}=3$, $\lambda_{\text{feat}}=1$, and $\lambda_{\text{depth}}=3$. Further details are provided in the appendix.

\subsection{Experimental Setup}
\label{sec:exp:setup}

\noindent\textbf{Simulation Benchmarks.}
We evaluate generalization across distinct axes using two simulation benchmarks. Specifically, we train our policy on \textbf{LIBERO}~\citep{liu2023libero}, a lifelong single-arm manipulation benchmark spanning diverse spatial layouts, object identities, and task goals. To rigorously assess out-of-distribution robustness, we then evaluate the trained models in a zero-shot manner on \textbf{LIBERO-Plus}~\citep{fei2025libero}, which introduces controlled environmental perturbations across dimensions such as camera viewpoint, lighting, and backgrounds. We report additional results in the appendix.

\definecolor{rankone}{RGB}{255,220,220}   
\definecolor{ranktwo}{RGB}{255,235,200}   
\definecolor{rankthree}{RGB}{255,250,200} 

\newcommand{\drop}[1]{{\scriptsize\,($\downarrow$#1)}}

\begin{table*}[t]
\centering

\caption{\textbf{Evaluation results on LIBERO and LIBERO-Plus.} Success rates are reported in \% with absolute performance drops from LIBERO to LIBERO-Plus shown in parentheses. Color highlights denote the top three performing methods within each column: \colorbox{rankone}{first}, \colorbox{ranktwo}{second}, and \colorbox{rankthree}{third}.}
\label{tab:sim}

\resizebox{\textwidth}{!}{%
\begin{tabular}{@{}l|c|cc|rrrrrrr}
\toprule
Method & Size & Orig. & Plus & Cam. & Robot & Lang. & Light & BG & Noise & Layout \\
\midrule

\multicolumn{11}{c}{\textit{VLAs}} \\
\midrule
$\pi_{0.5}$~\citep{pi05} & 3.3B
& 96.9
& \cellcolor{ranktwo}84.6\drop{12.3}
& \cellcolor{rankthree}72.0
& \cellcolor{rankone}76.6
& \cellcolor{ranktwo}86.5
& \cellcolor{rankthree}96.1
& \cellcolor{rankone}95.2
& \cellcolor{rankthree}86.7
& \cellcolor{rankone}86.0 \\

OpenVLA-OFT~\cite{openvla_oft} & 7B
& 97.1 & 69.6\drop{27.5}
& 56.4 & 31.9 & 79.5 & 88.7
& \cellcolor{rankthree}93.3
& 75.8 & 74.3 \\

RIPT-VLA~\cite{riptvla} & 7B
& 97.5 & 68.4\drop{29.1}
& 55.2 & 31.2 & 77.6 & 88.4
& 91.6
& 73.5 & 74.2 \\

$\pi_0$~\cite{wang2025pi} & 3.3B
& 91.3 & 69.3\drop{22.0}
& 61.0 & 40.8 & 63.7 & 89.3 & 84.1 & 80.1 & 75.9 \\

$\pi_0$-FAST~\cite{pi0_fast} & 3.3B
& 85.5 & 61.6\drop{23.9}
& 65.1 & 21.6 & 61.0 & 73.2 & 73.3 & 74.4 & 68.8 \\

UniVLA~\cite{univla} & 8.5B
& 95.2 & 42.9\drop{52.3}
& 1.8 & 46.2 & 69.5 & 69.0 & 81.0 & 21.2 & 31.9 \\

NORA~\cite{nora} & 3B
& 87.9 & 39.0\drop{48.9}
& 2.2 & 37.0 & 65.1 & 45.7 & 58.6 & 12.8 & 62.1 \\

OpenVLA~\citep{kim2024openvla} & 7B
& 76.5 & 15.6\drop{60.9}
& 0.8 & 3.5 & 23.0 & 8.1 & 34.8 & 15.2 & 28.5 \\

\midrule
\multicolumn{11}{c}{\textit{WAMs}} \\
\midrule
Cosmos-Policy~\citep{kim2026cosmos} & 2B
& \cellcolor{rankone}98.5
& \cellcolor{rankthree}82.4\drop{16.1}
& \cellcolor{ranktwo}73.4
& \cellcolor{rankthree}63.3
& \cellcolor{rankone}89.3
& \cellcolor{rankone}98.9
& 83.5
& \cellcolor{ranktwo}89.3
& \cellcolor{ranktwo}84.0 \\

Fast-WAM~\citep{yuan2026fast} & 6B
& \cellcolor{ranktwo}97.6 & 50.0\drop{47.5}
& 16.4 & 44.5 & 68.9 & 78.2 & 53.7 & 37.7 & 60.7 \\

WorldVLA~\cite{worldvla} & 7B
& 79.1 & 25.0\drop{54.1}
& 0.1 & 27.9 & 41.6 & 43.7 & 17.1 & 11.0 & 38.0 \\

\midrule
\multicolumn{11}{c}{\textit{Geometry-aware VLAs}} \\
\midrule
$\pi_{0.5}$ + Spatial Forcing~\cite{li2025spatial} & 3.3B
& 94.0 & 25.7\drop{58.3}
& 0.1 & 0.3 & 26.8 & 66.0 & 45.9 & 0.1 & 59.8 \\

$\pi_{0.5}$ + ROCKET~\cite{sun2026rocket} & 3.3B
& 95.3 & 47.5\drop{46.6}
& 30.9
& 75.6
& 29.3 & 69.2 & 47.0 & 25.4 & 62.0 \\

\midrule
\multicolumn{11}{c}{\textit{GAM}} \\
\midrule
\textbf{GAM (Ours)} & 1.4B
& \cellcolor{ranktwo}97.6
& \cellcolor{rankone}85.5\drop{12.1}
& \cellcolor{rankone}83.1
& \cellcolor{ranktwo}70.0
& \cellcolor{rankthree}84.8
& \cellcolor{ranktwo}97.2
& \cellcolor{ranktwo}94.3
& \cellcolor{rankone}95.3
& \cellcolor{rankthree}79.1 \\
\bottomrule
\end{tabular}%
}

\vspace{-10pt}
\end{table*}

\noindent\textbf{Real-Robot Setup.}
We train on four manipulation tasks ($\sim$200 demonstrations each) using wrist-mounted and third-person cameras, adhering to the simulation protocol. Since ground-truth geometry is unavailable in the real world, target future depth maps are obtained as pseudo-labels directly from the pretrained backbone GFM. We evaluate robustness via 20 trials per task, divided equally between nominal setups and perturbed environments, specifically varying external camera positions. See the appendix for robot environment with full task and evaluation details.

\noindent\textbf{Baselines.}
We compare GAM against representative baselines from three families discussed in §\ref{sec:related}: \textbf{VLAs}~\citep{kim2024openvla, openvla_oft, black2024pi_0, pi05, pi0_fast, nora, univla, riptvla}, \textbf{WAMs}~\citep{kim2026cosmos, worldvla, yuan2026fast}, and \textbf{geometry-aware VLAs}~\citep{li2025spatial, sun2026rocket}. For the real-robot setup, we compare against $\pi_{0.5}$~\cite{pi05} and Spatial Forcing~\cite{li2025spatial}. For fairness, comparisons utilize a matched evaluation protocol, with performance numbers either re-evaluated using available checkpoints or taken directly from their respective published benchmarks.

\subsection{Main Results}
\label{sec:exp:main}

\noindent\textbf{Simulation Results.} As shown in Table~\ref{tab:sim}, GAM achieves highly competitive success rates on the standard LIBERO benchmark, where performance is heavily saturated. Crucially, on the more challenging LIBERO-Plus benchmark, our model consistently outperforms competing baselines, demonstrating a remarkable improvement in the \textbf{camera-perturbation setting ($\uparrow$9.7\%p)}. This gain highlights the advantage of our end-to-end integration of the GFM. While existing geometry-aware VLAs only partially exploit GFM representations, GAM embeds the GFM throughout its entire predictive pathway to yield a deeply geometry-aware policy.

\begin{figure*}[t]
    \centering
    \includegraphics[width=\textwidth]{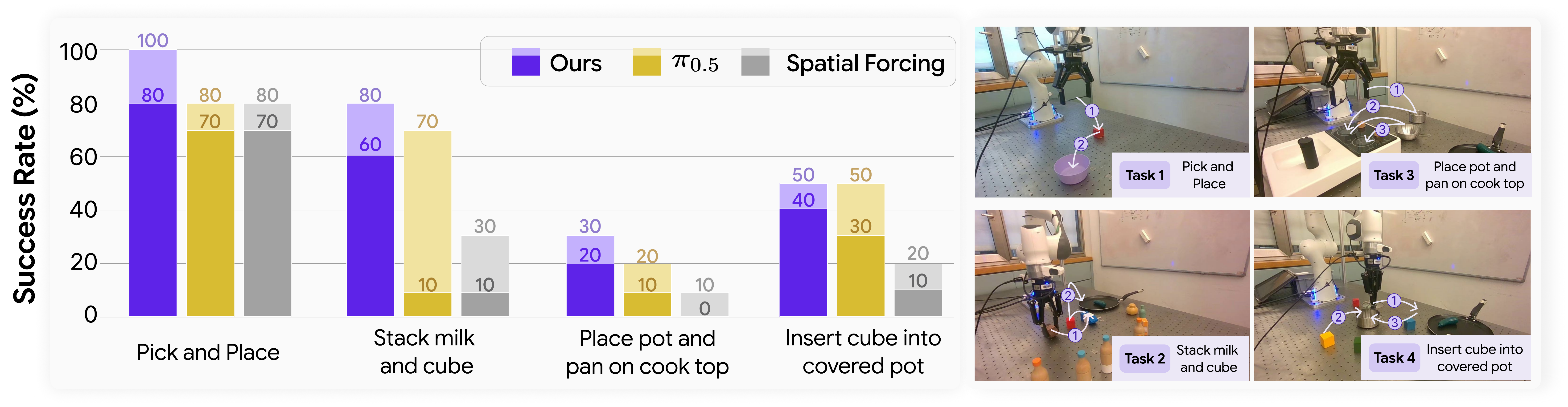}
    \vspace{-15pt}
    \caption{
    \textbf{Real-world robot tasks and results.}
    Each task is evaluated under both in-domain (Light bar) and out-of-domain (Dark bar) settings. The illustration of each task is shown on the right.
    }
    \label{fig:real_world_eval}
    \vspace{-10pt}
\end{figure*}

\noindent\textbf{Real-world Results.} 
To examine whether the gains observed in simulation transfer to physical execution, we additionally evaluate GAM in a real-world setting. Figure~\ref{fig:real_world_eval} shows that GAM substantially outperforms all baselines. In particular, our model remains robust under out-of-domain conditions (the camera-perturbation setting) where other baselines struggle. These results demonstrate that GAM generalizes to the real-world domain and is robust under perturbations, owing to its thorough exploitation of the GFM when training the policy.

\subsection{Ablation Study}
\label{sec:exp:ablation}

\paragraph{Post-training Component Analysis.}
Table~\ref{tab:loss_h_ablation} summarizes ablation study of key post-training components on Object suite of LIBERO and LIBERO-Plus. Pretraining is crucial for robustness: omitting it mildly affects nominal LIBERO but severely degrades LIBERO-Plus. With a pretrained backbone, removing $L_{\text{depth}}$ or $L_{\text{feat}}$ has minimal impact, suggesting geometric dynamics are already encoded.
Notably, even without pretraining, these future-prediction losses provide strong geometric supervision and substantially improve robustness on LIBERO-Plus. Finally, the horizon ablation shows that $H=1$ is sufficient and more robust than longer histories, consistent with prior observations that extended context can introduce spurious correlations~\cite{wen2020fighting,de2019causal}.

\newcommand{\cmark}{\textcolor{green!60!black}{\ding{51}}}
\renewcommand{\xmark}{\textcolor{red!75!black}{\ding{55}}}

\begin{wraptable}[12]{r}{0.66\textwidth}
\centering
\small
\renewcommand{\arraystretch}{0.96}
\vspace{-0.3em} 

\begin{minipage}[t]{0.56\linewidth}
\centering
\captionof{table}{\textbf{Component ablation.}}
\label{tab:loss_h_ablation}
\vspace{-0.55em}
\resizebox{1.02\linewidth}{!}{
\begin{tabular}{@{}cccccc@{}}
\toprule
Pretrain & $\mathcal{L}_{\text{depth}}$ & $\mathcal{L}_{\text{feat}}$ & H &
\begin{tabular}{@{}c@{}}Orig.\\SR (\%)\end{tabular} &
\begin{tabular}{@{}c@{}}Plus\\SR (\%)\end{tabular} \\
\midrule
\cmark & \cmark & \cmark & 1 & \textbf{99.6} & \textbf{89.7}\\
\midrule
\cmark & \cmark & \cmark & 2 & 97.2 & 84.4 \\
\cmark & \cmark & \cmark & 4 & 98.2 & 85.1 \\
\midrule
\cmark & \xmark & \cmark & 1 & 98.4 & 89.0 \\
\cmark & \xmark & \xmark & 1 & 98.6 & 89.5 \\
\cmark & \cmark & \xmark & 1 & \textbf{99.6} & \textbf{89.7} \\
\midrule
\xmark & \cmark & \cmark & 1 & 98.4 & 73.4 \\
\xmark & \xmark & \cmark & 1 & 95.2 & 66.5 \\
\xmark & \cmark & \xmark & 1 & 96.4 & 80.0 \\
\xmark & \xmark & \xmark & 1 & 93.6 & 50.0 \\
\bottomrule
\end{tabular}%
}
\end{minipage}%
\hfill
\begin{minipage}[t]{0.41\linewidth}
\centering

\captionof{table}{\textbf{Layer ablation.}}
\label{tab:predictor_layer_ablation}
\vspace{-0.65em}
{
\setlength{\tabcolsep}{2.7pt}%
\resizebox{0.85\linewidth}{!}{
\begin{tabular}{@{}ccc@{}}
\toprule
Split layer $L_s$  & Orig. (\%)& Plus (\%) \\
\midrule
0  & 5.4 & 1.8 \\
12 & \textbf{99.6} & \textbf{70.1} \\
19 & 95.6 & 63.4 \\
27 & 1.2 & 1.6 \\
33 & 0.0 & 0.0 \\
39 & 0.0 & 0.0 \\
\bottomrule
\end{tabular}%
}}

\vspace{-0.25em} 

\captionof{table}{\textbf{Inference cost.}}
\label{tab:inference_time}
\vspace{0.15em}
\resizebox{0.95\linewidth}{!}{%
\begin{tabular}{@{}lcc@{}}
\toprule
Method & Size & Time \\
\midrule
OpenVLA-OFT~\cite{kim2024openvla}     & 7B   & 77.8ms \\
$\pi_{0.5}$~\cite{pi05}               & 3.3B & 29.2ms   \\
Cosmos-Policy~\cite{kim2026cosmos}    & 2B   & 382.4ms  \\
\textbf{GAM (Ours)}                                  & \textbf{1.4B} & \textbf{6.9ms} \\
\bottomrule
\end{tabular}%
    }
\end{minipage}

\vspace{-1em} 
\end{wraptable}

\noindent\textbf{Split Layer $L_s$ Selection.}
Table~\ref{tab:predictor_layer_ablation} evaluates the depth of future predictor by shifting the split layer $L_s$ and re-initializing the predictor. We exclude future-depth loss in this experiment because it is not equally applicable to all split layers and could isolate the effect of $\mathcal{L}_{\text{feat}}$ itself. Our default choice of $L_s=12$ achieves peak performance, validating it as the optimal seam between frame-wise and cross-view attention. While layer 19 remains competitive, inserting the predictor too early ($L_s=0$) or late ($L_s\in\{27,33,39\}$) causes total performance collapse. This confirms that forecasted tokens require sufficient interaction through deep layers to properly integrate into the pretrained 3D geometric prior.

\subsection{Analysis}
\label{sec:exp:analysis}

\begin{wrapfigure}[11]{r}{0.32\textwidth}
    \centering
    \vspace{-18pt}
    \includegraphics[width=0.9\linewidth]{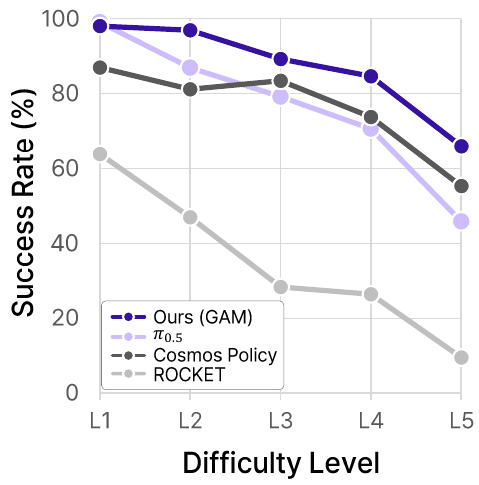}
    \vspace{-5pt}
    \caption{\textbf{Success rate vs.\ camera perturbation difficulty.}}
    \label{fig:camera_perturb}
\end{wrapfigure}
\textbf{Inference Speed and Model Size.}
As shown in Table~\ref{tab:inference_time}, GAM achieves the lowest latency among all baselines, requiring only \textbf{6.9\,ms ($\approx$145\,Hz)} for a single feed-forward pass and running up to 55$\times$ faster than the diffusion-based Cosmos Policy. 
 
All methods are benchmarked under the same setup, with further details provided in the appendix. By utilizing single-pass prediction, GAM avoids the multi-step denoising of diffusion policies, achieving low latency while matching prior accuracy and robustness with only 1.4B parameters.

\noindent\textbf{Robustness to Viewpoint and Scene Variation.} Figure~\ref{fig:camera_perturb} further breaks down the camera-perturbation results by difficulty level of LIBERO-Plus~\cite{fei2025libero}. GAM achieves consistently higher success rates than all baselines at every level, and the advantage remains clear even under the strongest perturbations.

\section{Conclusion and Limitation}
\label{sec:conclusion}
We introduced Geometric Action Model, which unifies geometry and action prediction with temporal world modeling inside a single shared GFM. By inserting a causal transformer between the GFM's shallow and deep layers, GAM autoregressively decodes actions and future geometries, resolving the spatial ambiguities of traditional foundation-model substrates. Across extensive simulation and real-world benchmarks, GAM achieves superior accuracy, faster inference, and strong out-of-distribution robustness to environmental perturbations.
The framework also has limitations. Its language reasoning and commonsense capabilities are bounded by the frozen text encoder; integrating a large language model or an external reasoning module is a natural next step.




\acknowledgments{This work was supported under project ID a144 as part of
the Swiss AI Initiative, through a grant from the ETH Domain and computational
resources provided by the Swiss National Supercomputing Centre (CSCS) under
the Alps infrastructure.}



\clearpage

\title{Geometric Action Model for Visuomotor Control\\\texttt{- Supplementary Materials -}}

\appendix

\setcounter{section}{0}
\renewcommand{\thesection}{\Alph{section}}


\section*{\Large Appendix}
This appendix provides experimental details, results, and analyses that complement the main paper. 
\begin{itemize}
\item Section~\ref{sec:appendix-experimental_settings} describes the training data, implementation details, simulation and real-world evaluation settings, baseline settings, and inference benchmark protocol.
\item Section~\ref{sec:appendix-additional_results} presents additional simulation benchmark results on LIBERO, LIBERO-Plus, and RoboCasa.
\item Section~\ref{sec:appendix-ablation_analysis} provides additional ablations and analyses, including backbone variants, pre-training ablations, split-layer analysis, action-token attention, and robustness trends.
\end{itemize}

\label{secA}

\section{Experimental Settings and Reproducibility Details}
\label{sec:appendix-experimental_settings}

GAM is trained in two stages, following standard practice for generalist robot policies~\citep{openvla_oft, black2024pi_0}. The first stage jointly trains the predictor, action head, and the GFM backbone end-to-end on a large mixture of single-arm robot data. The model was trained using 64 NVIDIA GH200 GPUs witch batch size of 1024, which takes approximately \textasciitilde 96 hours. The second stage fine-tunes the entire model on each benchmark's official training set before evaluation. The second stage on simulation benchmark was trained using 16 NVIDIA GH200 GPUs with batch size of 160, which takes \textasciitilde 48 hours.
\subsection{Pre-training Details}
\label{sec:appendix-pretraining_details}
\begin{figure*}[h]
    \centering
    \includegraphics[width=\textwidth]{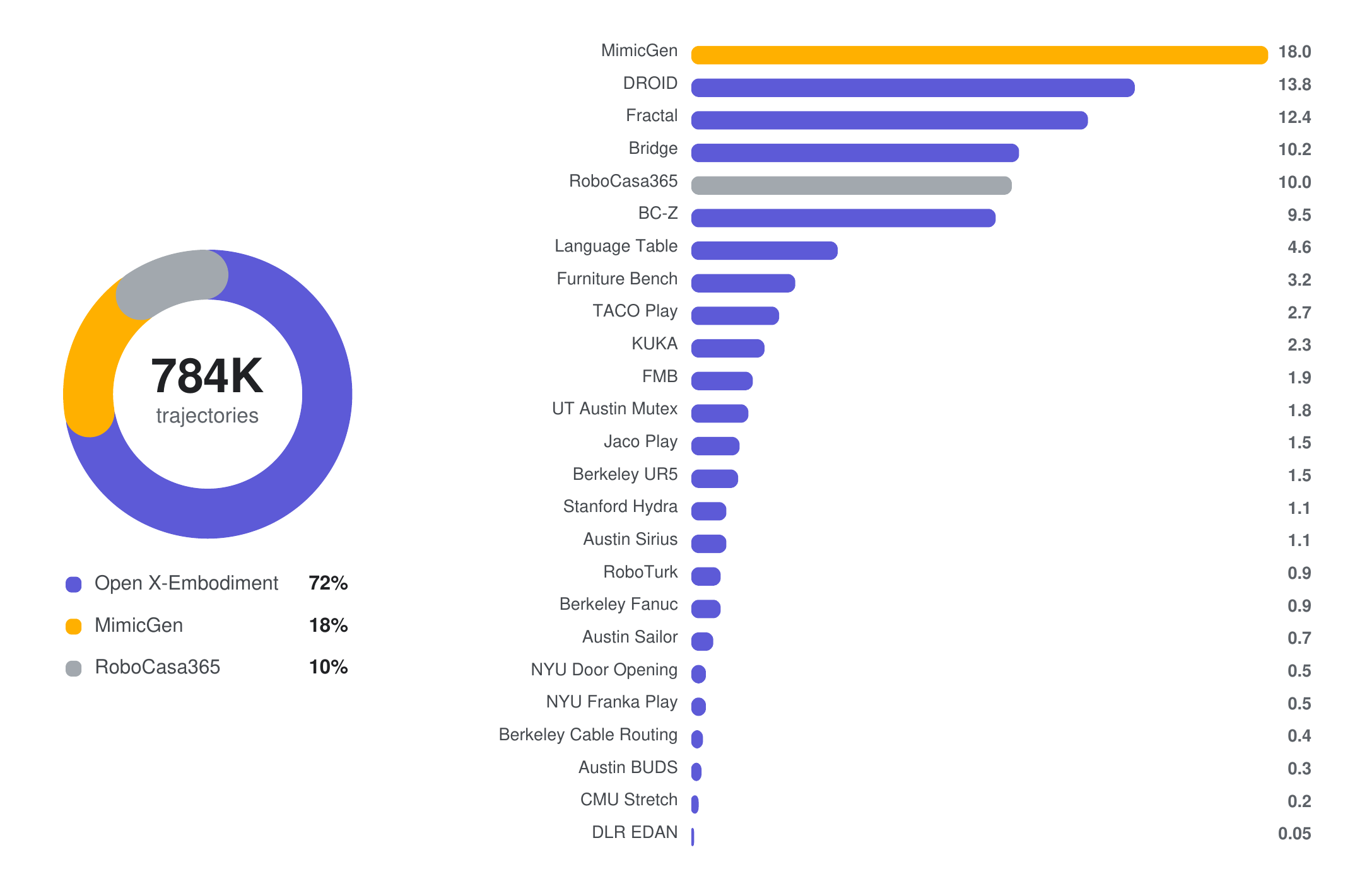}
    \caption{
    \textbf{Training Dataset Mixture.} We illustrate the dataset mixture utilized during pretraining, detailing the relative proportions of each constituent dataset. The pie chart shows the high-level source mixture, and the bar chart shows the percentage of each constituent dataset relative to the entire training corpus.
    }
    \label{supfig:training_data}
\end{figure*}

We pre-train GAM on a weighted mixture of Open-X Embodiment~\cite{collaboration2023open} (OXE), MimicGen~\cite{mandlekar2023mimicgen}, and RoboCasa365~\cite{nasiriany2026robocasa365}. OXE provides broad real-robot coverage across multiple embodiments and manipulation domains, while MimicGen and RoboCasa365 provide simulation demonstrations with clean geometric supervision. The sampling ratios are 72\%, 18\%, and 10\% for OXE, MimicGen, and RoboCasa365, respectively. For future-depth supervision, we use teacher pseudo-depth for OXE and re-rendered simulator depth for MimicGen and RoboCasa365. Figure~\ref{supfig:training_data} shows the dataset mixture used for pre-training.

For OXE, we use the subset whose actions can be mapped to our common control interface and exclude datasets that are incompatible with our action space.. For RoboCasa365, we use only the manipulation-task subset. Across all three sources, we keep the original task language provided by the datasets and do not synthesize additional instructions. The language encoder is kept frozen during pre-training.

All datasets are converted to a common observation and action format before training. Images are resized to \(224\times224\), and the model uses two RGB views when available: an external view and a wrist view. Following Cosmos-policy~\cite{kim2026cosmos} and $\pi_{0.5}$~\cite{pi05}, standard image augmentations such as random cropping, rotation, and color jitter are applied during training and disabled during evaluation.

\begin{table*}[t]
\centering
\caption{\textbf{Hyperparameters for LIBERO, LIBERO-Plus and real-world experiments.}}
\label{suptab:gam_hp}
\resizebox{1.0\linewidth}{!}{
\begin{tabular}{ll}
\toprule
hyperparameter & value \\
\midrule
\# GPUs & 8 $\times$ NVIDIA GH200 \\
learning rate (LR) & 5.16e-5 backbone; 5.16e-4 action head and predictor \\
total batch size & 160 \\
input images & 1 external camera image, 1 wrist-mounted camera image \\
input image size & 224 x 224 px \\
use observation history & no (use single-step inputs) \\
action chunk size & 8 steps (predict 8, execute all 8 open-loop at test time) \\
use proprio (robot state) & yes \\
\# trainable parameters & 983.2M total \\
image augmentations & 90\% random crop, $\pm5^\circ$ rotation, color jitter, JPEG q95: \\
& \;\; \texttt{random\_resized\_crop=dict(scale=[0.9, 0.9], ratio=[1.0, 1.0])} \\
& \;\; \texttt{base\_only\_rotation=[$\pm$5 deg]} \\
& \;\; \texttt{random\_brightness=[0.3]} \\
& \;\; \texttt{random\_contrast=[0.6, 1.4]} \\
& \;\; \texttt{random\_saturation=[0.5, 1.5]} \\
& \;\; \texttt{random\_hue=[0.05]} \\
& \;\; \texttt{jpeg\_quality=95} \\
\bottomrule
\end{tabular}}
\end{table*}

\subsection{Simulation Experiments Details}
\label{sec:appendix-simulation_details}
We adopt the LIBERO evaluation protocol established by OpenVLA~\cite{kim2024openvla} and OpenVLA-OFT~\cite{openvla_oft}. Specifically, we evaluate on the four standard LIBERO task suites: LIBERO-Spatial, LIBERO-Object, LIBERO-Goal, and LIBERO-Long, each consisting of 10 tasks. Following OpenVLA-OFT, we train on filtered LIBERO demonstrations by removing unsuccessful episodes and filtering idle/no-op frames, i.e., training samples with near-zero actions. We fine-tune a separate policy for each LIBERO suite. 

We report task execution success rate (SR, \%) as our primary evaluation metric. For the original LIBERO benchmark, each task is evaluated over 50 randomized trials, resulting in 500 rollouts per suite. For LIBERO-Plus, we follow the official evaluation setting and use one rollout per perturbed task instance. All models are trained with a global batch size of 160 for up to 110k training steps, until convergence is achieved. 

For baseline comparison, we re-evaluate $\pi_{0.5}$ and Cosmos-Policy under our evaluation setting using publicly available checkpoints. We also re-evaluate $\pi_{0.5}$ + Spatial Forcing and $\pi_{0.5}$ + ROCKET using reproduced checkpoints from ROCKET~\cite{sun2026rocket}. Results for the remaining baselines are taken from~\citet{fei2025libero} and ~\citet{zheng2026pokevla}.

\subsection{Real-World Experiments Details}
\label{sec:appendix-real_world_details}
\begin{figure*}[t]
    \centering
    \includegraphics[width=\textwidth]{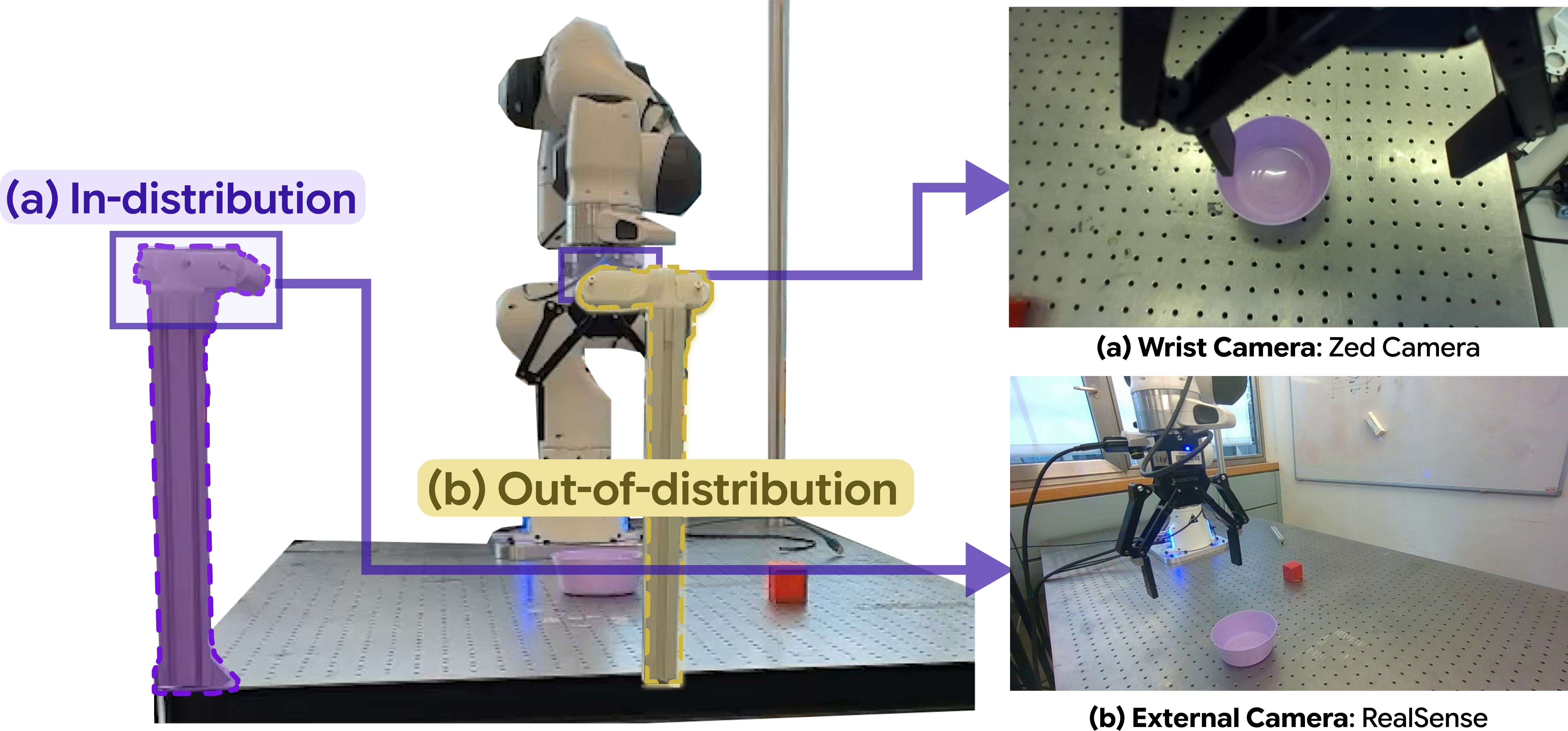}
    \caption{
    \textbf{Real-world experiments environment setup and ID vs. OOD Camera setup.}
    }
    \label{supfig:real_world_setup}
\end{figure*}

Figure~\ref{supfig:real_world_setup} illustrates the experimental environment used for our real-world evaluations. Our hardware setup includes a wrist-mounted ZED Camera and an external RealSense camera providing a third-person perspective. 

\begin{figure*}[t]
    \centering
    \includegraphics[width=\textwidth]{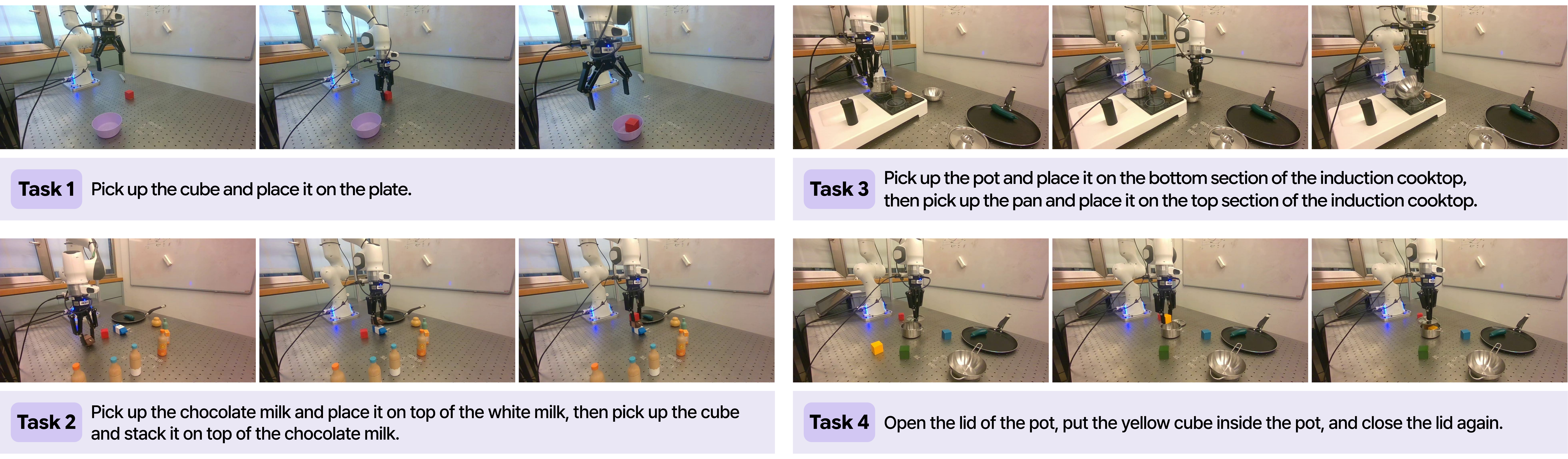}
    \caption{
    \textbf{Illustration of four real-world manipulation tasks.}
    }
    \label{supfig:real_world_tasks}
\end{figure*}

For training and evaluation, we defined four distinct tasks: Pick and place, Stack milk and cube, Place pot and pan on cooktop, and Insert cube into covered pot. We collected teleoperated demonstrations for each task: 284, 202, 184, and 169 demonstrations, respectively. Figure~\ref{supfig:real_world_tasks} shows the text instructions and corresponding visual illustrations for each task. All tasks were jointly trained within a unified dataset. The hardware specifications and hyperparameters used to train GAM, alongside the baselines ($\pi_{0.5}$~\cite{pi05} and Spatial Forcing~\cite{li2025spatial}), are detailed in Tables~\ref{suptab:gam_hp}, \ref{suptab:pi_hp}, and \ref{suptab:sf_hp}, respectively. All baseline models were trained until convergence using the default hyperparameters from the paper.

During evaluation, we measured the success rate of each task across 20 trials. Specifically, 10 trials were conducted under a normal setup (ID), while the remaining 10 trials evaluated robustness under an out-of-distribution (OOD) setup with camera perturbation. The camera perturbation was introduced by applying a translation of 85~cm and a rotation of 45$^\circ$ to the external camera shown in Figure~\ref{supfig:real_world_setup}. The left side of Figure~\ref{supfig:real_world_setup}, we provide a visualization comparing ID and OOD environment settings. This perturbation setup was kept consistent across all evaluations.

\begin{table*}[t]
\centering
\label{suptab:pi_sf_hp}
\scriptsize
\begin{minipage}[t]{0.49\linewidth}
\centering
\captionof{table}{\textbf{$\mathbf{\pi_{0.5}}$ hyperparameters.}}
\label{suptab:pi_hp}
\resizebox{\linewidth}{!}{
\begin{tabular}{ll}
\toprule
hyperparameter & value \\
\midrule
\# GPUs & 8 x NVIDIA H100 GPU \\
learning rate (LR) & 5e-5 \\
total batch size & 16 \\
input images & 1 external camera image, 1 wrist-mounted camera image \\
input image size & 224 x 224 px \\
use observation history & no (use single-step inputs) \\
action chunk size & 10 steps (predict 10, execute all 10 open-loop at test time) \\
use proprio (robot state) & yes \\
\# trainable parameters & 3.3B total \\
diffusion sampling algorithm & flow matching \\
number of integration steps & 10 \\
image augmentations & 90\% random crops, color jitter: \\
& \;\; \makecell[l]{\texttt{random\_resized\_crop=dict(scale=[0.9, 0.9],}\\ \texttt{ratio=[1.0, 1.0])}} \\
\bottomrule
\end{tabular}}
\end{minipage}
\hfill
\begin{minipage}[t]{0.49\linewidth}
\centering
\captionof{table}{\textbf{Spatial Forcing hyperparameters.}}
\label{suptab:sf_hp}
\resizebox{\linewidth}{!}{
\begin{tabular}{ll}
\toprule
hyperparameter & value \\
\midrule
\# GPUs & 8 x NVIDIA H100 GPU \\
learning rate (LR) & 2.5e-5 \\
total batch size & 16 \\
input images & 1 external camera image, 1 wrist-mounted camera image \\
input image size & 224 x 224 px \\
use observation history & no (use single-step inputs) \\
action chunk size & 10 steps (predict 10, execute all 10 open-loop at test time) \\
use proprio (robot state) & yes \\
\# trainable parameters & 853M total \\
image augmentations & 90\% random crops, color jitter: \\
& \;\; \makecell[l]{\texttt{random\_resized\_crop=dict(scale=[0.9, 0.9],}\\ \texttt{ratio=[1.0, 1.0])}} \\
\bottomrule
\end{tabular}}
\end{minipage}
\end{table*}

\subsection{Real-World Experiments Baseline Training Details}
\label{sec:appendix-baseline_training}

For the baseline methods of real-world experiments, we follow the training recipes and default hyperparameters from the corresponding papers~\cite{li2025spatial, pi05}, changing only the task data and camera streams to match our evaluation setup (Table~\ref{suptab:pi_hp} and \ref{suptab:sf_hp}). Both $\pi_{0.5}$~\cite{pi05} and Spatial Forcing~\cite{li2025spatial} use the same two RGB inputs as GAM, one external camera and one wrist-mounted camera, resized to \(224\times224\). We keep each baseline's original inference protocol to preserve its intended deployment behavior. The main baseline-specific settings are summarized below; $\pi_{0.5}$ uses flow-matching action decoding with 10 integration steps, while Spatial Forcing adopts its feature alignment loss recipe.

\subsection{Inference Latency Comparison Details}
\label{sec:appendix-inference_details}

\renewcommand{\cmark}{\textcolor{green!60!black}{\ding{51}}}
\renewcommand{\xmark}{\textcolor{red!75!black}{\ding{55}}}

\begin{table}[t]
\centering
\small
\caption{
\textbf{Model-only inference latency on a single GH200 GPU.}
}
\setlength{\tabcolsep}{4.8pt}
\begin{tabular}{@{}lccccc@{}}
\toprule
\textbf{Policy} &
\begin{tabular}{@{}c@{}}\textbf{Official}\\\textbf{PyTorch}\end{tabular} &
\begin{tabular}{@{}c@{}}\textbf{bf16}\\\textbf{precision}\end{tabular} &
\begin{tabular}{@{}c@{}}\textbf{Torch}\\\textbf{Compile}\end{tabular} &
\begin{tabular}{@{}c@{}}\textbf{CUDA}\\\textbf{Graphs}\end{tabular} &
\begin{tabular}{@{}c@{}}\textbf{Model-only}\\\textbf{latency}\end{tabular} \\
\midrule
\textbf{GAM (Ours)}    & \cmark & \cmark & \cmark & \xmark  & 17.5\,ms \\
\textbf{GAM (Ours)} & \cmark & \cmark & \cmark & \cmark   & \textbf{6.9\,ms} \\
\midrule
pi0.5                 & \cmark & \cmark & \cmark & \xmark  & 29.2\,ms \\
OpenVLA-OFT           & \cmark & \cmark & \cmark & \xmark   & 70.1\,ms \\
Cosmos Policy         & \cmark & \cmark & \cmark & \xmark   & 382.4\,ms \\
\bottomrule
\end{tabular}

\label{tab:inference_time_supp}
\end{table}
Table~\ref{tab:inference_time_supp} reports the runtime configuration used for the inference-speed comparison in the main paper. 
All policies are evaluated on a single GH200 GPU with the same canonical observation input, bf16 precision, warmup and measurement protocol, and model-only latency metric, excluding model loading and input preprocessing. 

In main paper, We use the official PyTorch inference path for each baseline. For $\pi_{0.5}$, whose original implementation is based on JAX, we use the official PyTorch implementation. To separate common compiler and runtime effects from deployment-specific execution, we report a matched setting in which all policies use Torch Compile and none uses CUDA Graphs.
Under this setting, GAM requires 17.5\,ms for a single feed-forward action prediction, compared to 29.2\,ms for $\pi_{0.5}$, 70.1\,ms for OpenVLA-OFT, and 382.4\,ms for Cosmos Policy. 
GAM's deployment setting further uses CUDA Graphs over its static single-pass inference path, reducing latency to 6.9\,ms. This corresponds to approximately 145\,Hz control and up to a \textbf{55.4}$\times$ speedup over the diffusion-based Cosmos Policy.

\subsection{Model Size Breakdown}
\label{sec:appendix-params_breakdown}
\begin{table}[t]
\centering
\caption{\textbf{Parameter breakdown of the DA3-based GAM model.}}
\label{suptab:gam_param_breakdown}
\resizebox{1.0\linewidth}{!}{
\begin{tabular}{llll}
\toprule
Module & Parameters & Trainable? & Trainable parameters \\
\midrule
backbone (ViT-Giant, 40 blocks) & 1136.5M & blocks 13--39 trainable; blocks 0--12 frozen & $\approx$765M \\
DPT head & 50.1M & frozen & 0 \\
Causal Future Predictor & 210.2M & trainable & 210.2M \\
action head & 8.0M & trainable & 8.0M \\
\midrule
total & $\approx$1404.8M & -- & $\approx$983.2M \\
\bottomrule
\end{tabular}}
\end{table}
Table~\ref{suptab:gam_param_breakdown} details the parameter breakdown of the DA3-based GAM architecture. As reported, GAM uses a 1.4B-parameter model, making it substantially smaller than VLM-based and video-diffusion-based baselines such as $\pi_{0.5}$, OpenVLA-OFT, and Cosmos-Policy. This compact size comes from repurposing a pretrained geometric backbone~\cite{lin2025depth} as the shared substrate for perception, future-geometry prediction, and action decoding, rather than attaching a large language model or video-generation model as the policy backbone.

Of the full model, approximately 983.2M parameters are trainable. Most of these trainable parameters come from the later blocks of the ViT-Giant backbone, while the initial geometric layers and the DPT depth head remain frozen to preserve pretrained geometric structure. The Causal Future Predictor and lightweight action head are fully trainable and account for the remaining trainable parameters. This design allows GAM to adapt the geometric representation for control while keeping the overall model size below the larger foundation-model baselines in the main comparison.

\section{Additional Benchmark Results}
\label{sec:appendix-additional_results}
\subsection{Additional Results on RoboCasa-kitchen}
\label{sec:appendix-robocasa}
\newcommand{\tabitem}{\hspace{0.3em}$+$\hspace{0.25em}}

\begin{wraptable}[15]{r}{0.38\textwidth}
\vspace{-10pt}
\centering
\footnotesize
\setlength{\tabcolsep}{5.0pt}
\renewcommand{\arraystretch}{1.06}

\caption{\textbf{Average success rates on RoboCasa Kitchen.}}
\label{tab:training-demos-average-sr}

\begin{tabular}{@{}lc@{}}
\toprule
Method & Avg. SR (\%) \\
\midrule
GROOT-N1              & 49.6 \\
\tabitem DreamGen     & 57.6 \\
\tabitem DUST         & 58.5 \\
\addlinespace[0.25em]
UWM                   & 60.8 \\
$\pi_0$               & 62.5 \\
\addlinespace[0.25em]
GROOT-N1.5            & 64.1 \\
\tabitem HAMLET       & 66.4 \\
\addlinespace[0.25em]
Video Policy          & 66.0 \\
FLARE                 & 66.4 \\
Cosmos Policy         & 67.1 \\
\textbf{GAM (Ours)}  & \textbf{69.4} \\

\bottomrule
\end{tabular}

\vspace{-0.8em}
\end{wraptable}

RoboCasa-Kitchen is a simulation benchmark derived from RoboCasa~\cite{nasiriany2024robocasa}, which focuses on everyday manipulation in realistic and diverse kitchen environments. We evaluate on 24 kitchen manipulation tasks that cover pick-and-place, articulated-object interaction, appliance control, and coffee-related manipulation skills.

Because our base pre-training setup uses two camera views, whereas RoboCasa-Kitchen adopts a 3-view observation, we further train GAM from the base pre-training checkpoint for 3-view format. We also increase the action chunk size from 8 to 16 steps to better accommodate the longer-horizon nature of RoboCasa-Kitchen tasks. For the benchmark demonstrations, we re-extract depth for the 300 demonstrations per task and use only successful trajectories for training. As summarized in Table~\ref{tab:training-demos-average-sr}, We find that GAM outperforms existing baselines, including Cosmos Policy~\cite{kim2026cosmos}, on RoboCasa-Kitchen,  We additionally provide the per-task breakdown of RoboCasa-Kitchen in Table~\ref{tab:robocasa_taskwise_mean}.

\begin{table}[t]
    \centering
    \caption{
    \textbf{Task-wise success rates on RoboCasa-Kitchen.}
    We report per-task success rates and the overall average.
    }
    \label{tab:robocasa_taskwise_mean}
    \small
    \setlength{\tabcolsep}{5pt}
    \renewcommand{\arraystretch}{1.05}
    \resizebox{\linewidth}{!}{
    \begin{tabular}{lc lc lc lc}
        \toprule
        Task & SR & Task & SR & Task & SR & Task & SR \\
        \midrule
        PnPCabToCounter       & 40.0\% & PnPSinkToCounter   & 71.3\% & CloseDoubleDoor     & 90.0\% & TurnOffSinkFaucet & 88.3\% \\
        PnPCounterToCab       & 50.3\% & PnPStoveToCounter  & 56.3\% & CloseSingleDoor     & 100.0\% & TurnSinkSpout     & 84.7\% \\
        PnPCounterToMicrowave & 29.3\% & OpenSingleDoor     & 70.3\% & OpenDrawer          & 92.3\% & CoffeePressButton & 79.0\% \\
        PnPCounterToSink      & 75.7\% & OpenDoubleDoor     & 98.3\% & CloseDrawer         & 99.3\% & TurnOnMicrowave   & 90.7\% \\
        PnPCounterToStove     & 44.0\% & TurnOnStove        & 74.3\% & TurnOnSinkFaucet    & 89.7\% & TurnOffMicrowave  & 94.3\% \\
        PnPMicrowaveToCounter & 11.3\% & TurnOffStove       & 30.0\% & CoffeeServeMug      & 71.3\% & CoffeeSetupMug    & 33.7\% \\
        \midrule
        \multicolumn{7}{r}{\textbf{Overall}} & \textbf{69.4\%} \\
        \bottomrule
    \end{tabular}
    }
\end{table}

\subsection{Additional LIBERO and LIBERO-Plus Results}
\label{sec:appendix-additional_libero}

\definecolor{rankone}{RGB}{255,220,220}   
\definecolor{ranktwo}{RGB}{255,235,200}   
\definecolor{rankthree}{RGB}{255,250,200} 
\newcommand{\rkone}[1]{\cellcolor{rankone}#1}
\newcommand{\rktwo}[1]{\cellcolor{ranktwo}#1}
\newcommand{\rkthree}[1]{\cellcolor{rankthree}#1}

\begin{table*}[h]
\centering
\caption{\textbf{LIBERO-Plus robustness results across four task suites.} Success rates are reported in \%. "Orig." denotes the results of LIBERO benchmark without any perturbation. } 
\vspace{0.2em}
\label{tab:libero-plus-2x2}
\scriptsize
\setlength{\tabcolsep}{1.5pt}
\renewcommand{\arraystretch}{0.76}

\begin{minipage}[t]{0.49\textwidth}
\centering
\textbf{(a) Spatial}
\vspace{0.2em}
\resizebox{\linewidth}{!}{%
\begin{tabular}{@{}lrrrrrrrrr@{}}
\toprule
Method & Orig. & Cam. & Robot & Lang. & Light & BG & Noise & Layout & Total \\
\midrule
\textbf{GAM (Ours)} & \rktwo{98.6} & \rkone{91.5} & \rktwo{79.4} & \rkthree{95.6} & \rkone{100.0} & \rktwo{99.6} & \rkone{96.6} & 94.3 & \rkone{93.4} \\
$\pi_{0.5}$ & \rkone{98.8} & 76.6 & \rkone{86.3} & \rktwo{96.4} & 97.9 & \rktwo{99.6} & 92.0 & \rkone{98.2} & \rktwo{92.0} \\
Cosmos-policy & 98.1 & 83.5 & \rkthree{59.7} & \rktwo{96.4} & \rktwo{99.7} & 85.3 & 91.7 & \rkthree{95.1} & \rkthree{87.3} \\
Fast-WAM & \rkthree{98.2} & 14.4 & 44.0 & 69.5 & 87.3 & 69.8 & 35.0 & 60.5 & 54.4 \\
OpenVLA-OFT & 97.6 & \rktwo{88.3} & 40.0 & 80.5 & \rkthree{98.3} & 97.3 & \rktwo{96.3} & 93.9 & 84.0 \\
RIPT-VLA & 97.5 & 85.4 & 38.0 & \rkone{99.7} & \rktwo{99.7} & \rkone{100.0} & 92.0 & 92.3 & 85.8 \\
$\pi_0$ & 96.8 & 70.7 & 49.1 & 67.9 & 92.8 & 95.0 & 87.7 & 94.0 & 78.6 \\
$\pi_0^{\ast}$ & 96.8 & 17.8 & 6.6 & 58.8 & 89.7 & 90.7 & 90.9 & 89.1 & 60.7 \\
$\pi_0$-Fast & 96.4 & \rkthree{87.2} & 26.9 & 84.2 & 37.0 & \rkthree{97.7} & \rkthree{93.2} & \rktwo{95.5} & 74.4 \\
UniVLA & 96.5 & 1.1 & 52.6 & 83.9 & 96.6 & 90.7 & 15.7 & 69.5 & 55.5 \\
NORA & 92.2 & 4.3 & 50.9 & 63.8 & 66.8 & 65.5 & 12.5 & 84.6 & 47.6 \\
WorldVLA & 85.6 & 0.0 & 44.3 & 46.3 & 65.1 & 19.8 & 11.7 & 46.1 & 32.5 \\
OpenVLA & 84.7 & 0.0 & 3.7 & 27.7 & 12.3 & 50.4 & 12.0 & 40.7 & 19.4 \\
$\pi_{0.5}$ + ROCKET & 96.4 & 14.9 & 18.0 & 35.4 & 50.7 & 45.0 & 13.7 & 48.6 & 31.5 \\
\bottomrule
\end{tabular}%
}
\end{minipage}
\hfill
\begin{minipage}[t]{0.49\textwidth}
\centering
\textbf{(b) Object}
\vspace{0.2em}
\resizebox{\linewidth}{!}{%
\begin{tabular}{@{}lrrrrrrrrr@{}}
\toprule
Method & Orig. & Cam. & Robot & Lang. & Light & BG & Noise & Layout & Total \\
\midrule
\textbf{GAM (Ours)} & \rktwo{99.6} & \rkone{91.4} & \rkone{76.9} & \rkone{100.0} & \rkone{100.0} & \rkone{99.7} & \rkone{99.2} & 79.7 & \rkone{90.6} \\
$\pi_{0.5}$ & 98.2 & \rkthree{86.4} & \rktwo{71.9} & 91.0 & \rkthree{99.0} & \rktwo{99.2} & \rkthree{96.2} & \rkone{91.1} & \rktwo{89.9} \\
Cosmos-policy & \rkone{100.0} & \rktwo{88.6} & 61.6 & 94.1 & \rktwo{99.7} & 96.8 & \rktwo{97.4} & \rktwo{86.4} & \rkthree{88.3} \\
Fast-WAM & \rkone{100.0} & 25.3 & \rkthree{64.1} & \rkthree{96.3} & 97.9 & 77.8 & 63.7 & 73.0 & 71.2 \\
OpenVLA-OFT & 98.4 & 38.9 & 25.4 & \rktwo{99.0} & 73.7 & \rkthree{97.6} & 72.3 & 71.8 & 66.5 \\
RIPT-VLA & 97.5 & 37.9 & 26.4 & 80.8 & 85.9 & \rktwo{99.2} & 68.0 & 70.1 & 64.3 \\
$\pi_0$ & \rkthree{98.8} & 80.1 & 31.9 & 75.4 & 94.3 & 85.9 & 87.9 & 76.2 & 74.7 \\
$\pi_0^{\ast}$ & \rkthree{98.8} & 22.2 & 8.3 & 70.0 & 90.9 & 91.1 & 87.0 & 76.2 & 61.4 \\
$\pi_0$-Fast & 96.8 & 72.0 & 27.6 & 71.5 & 71.0 & 95.2 & 93.1 & \rkthree{84.5} & 72.7 \\
UniVLA & 96.8 & 0.0 & 42.2 & 86.9 & 25.6 & 81.5 & 10.4 & 27.3 & 36.7 \\
NORA & 95.4 & 0.5 & 28.4 & 76.4 & 25.3 & 54.8 & 5.7 & 55.8 & 34.4 \\
WorldVLA & 89.0 & 0.0 & 26.4 & 57.2 & 20.5 & 17.3 & 18.0 & 53.6 & 28.6 \\
OpenVLA & 88.4 & 0.5 & 4.5 & 21.0 & 1.0 & 45.2 & 11.4 & 22.4 & 14.0 \\
$\pi_{0.5}$ + ROCKET & \rkthree{98.8} & 41.4 & 27.1 & 44.6 & 91.6 & 69.8 & 34.1 & 83.6 & 53.9 \\
\bottomrule
\end{tabular}%
}
\end{minipage}

\vspace{0.8em}

\begin{minipage}[t]{0.49\textwidth}
\centering
\textbf{(c) Goal}
\vspace{0.2em}
\resizebox{\linewidth}{!}{%
\begin{tabular}{@{}lrrrrrrrrr@{}}
\toprule
Method & Orig. & Cam. & Robot & Lang. & Light & BG & Noise & Layout & Total \\
\midrule
\textbf{GAM (Ours)} & 97.4 & \rkone{94.9} & \rktwo{67.5} & \rkthree{67.8} & \rkone{100.0} & \rktwo{91.8} & \rkone{97.1} & \rkthree{64.5} & \rkone{80.4} \\
$\pi_{0.5}$ & \rktwo{98.0} & \rktwo{77.2} & \rkone{73.6} & \rktwo{70.5} & 93.2 & \rkone{92.5} & \rktwo{87.3} & \rkone{70.4} & \rktwo{79.3} \\
Cosmos-policy & \rkone{98.2} & 64.0 & \rkthree{64.1} & \rkone{73.4} & \rkone{100.0} & \rkthree{85.1} & 75.2 & \rktwo{65.2} & \rkthree{73.5} \\
Fast-WAM & 97.0 & 8.1 & 24.7 & 49.8 & 75.3 & 44.5 & 23.7 & 48.0 & 39.2 \\
OpenVLA-OFT & \rkthree{97.9} & 62.0 & 25.2 & 53.2 & \rkthree{93.9} & \rkone{92.5} & 75.2 & 59.1 & 63.0 \\
RIPT-VLA & 97.5 & 65.7 & 23.2 & 45.4 & 74.2 & 79.7 & 71.0 & 59.8 & 58.0 \\
$\pi_0$ & 95.8 & 56.6 & 43.3 & 43.2 & 90.3 & 84.7 & \rkthree{82.8} & 59.8 & 63.4 \\
$\pi_0^{\ast}$ & 95.8 & 12.3 & 5.6 & 39.3 & 84.2 & 76.5 & 76.5 & 44.7 & 44.9 \\
$\pi_0$-Fast & 88.6 & \rkthree{70.8} & 20.5 & 47.3 & \rktwo{95.3} & 60.9 & 69.7 & 51.6 & 57.5 \\
UniVLA & 95.6 & 3.9 & 37.9 & 45.6 & 89.6 & 78.3 & 33.5 & 22.6 & 40.7 \\
NORA & 89.4 & 2.9 & 31.1 & 56.6 & 60.6 & 60.5 & 18.2 & 53.9 & 38.8 \\
WorldVLA & 82.6 & 0.3 & 30.6 & 42.2 & 68.8 & 30.3 & 13.5 & 47.4 & 31.8 \\
OpenVLA & 79.2 & 2.5 & 2.7 & 21.5 & 9.0 & 27.1 & 19.5 & 25.6 & 15.1 \\
$\pi_{0.5}$ + ROCKET & 96.6 & 41.2 & 36.9 & 27.6 & 57.0 & 47.0 & 28.2 & 46.8 & 39.7 \\
\bottomrule
\end{tabular}%
}
\end{minipage}
\hfill
\begin{minipage}[t]{0.49\textwidth}
\centering
\textbf{(d) Long}
\vspace{0.2em}
\resizebox{\linewidth}{!}{%
\begin{tabular}{@{}lrrrrrrrrr@{}}
\toprule
Method & Orig. & Cam. & Robot & Lang. & Light & BG & Noise & Layout & Total \\
\midrule
\textbf{GAM (Ours)} & 94.6 & \rkone{62.5} & \rkthree{62.8} & \rkone{95.3} & 91.6 & \rkthree{88.2} & \rktwo{89.5} & \rkthree{79.5} & \rktwo{78.0} \\
$\pi_{0.5}$ & 92.4 & \rkthree{49.2} & \rkone{76.1} & \rkthree{89.6} & \rktwo{93.8} & \rkone{90.3} & \rkthree{73.1} & \rktwo{85.9} & \rkthree{77.9} \\
Cosmos-policy & \rkone{97.6} & \rktwo{58.9} & \rktwo{67.4} & \rktwo{94.8} & \rkone{96.4} & 68.9 & \rkone{91.8} & \rkone{92.9} & \rkone{81.0} \\
Fast-WAM & \rkthree{95.2} & 17.7 & 45.3 & 60.1 & 52.2 & 22.8 & 28.5 & 61.2 & 41.1 \\
OpenVLA-OFT & 94.5 & 38.7 & 38.2 & 87.0 & 89.4 & 86.8 & 63.5 & 76.9 & 66.4 \\
RIPT-VLA & \rktwo{97.5} & 34.1 & 38.4 & 88.3 & \rkthree{93.4} & \rktwo{89.3} & 66.4 & 79.2 & 67.5 \\
$\pi_0$ & 73.8 & 38.7 & 39.9 & 69.7 & 79.2 & 72.3 & 64.6 & 77.6 & 61.3 \\
$\pi_0^{\ast}$ & 85.2 & 3.8 & 3.6 & 68.4 & 74.5 & 69.5 & 64.4 & 69.6 & 48.4 \\
$\pi_0$-Fast & 60.2 & 33.2 & 12.0 & 43.6 & 91.6 & 44.6 & 46.1 & 47.8 & 43.4 \\
UniVLA & 92.0 & 1.9 & 53.2 & 64.2 & 65.7 & 74.4 & 25.4 & 16.4 & 39.9 \\
NORA & 74.6 & 1.2 & 39.4 & 64.0 & 30.3 & 54.0 & 15.1 & 59.5 & 36.3 \\
WorldVLA & 59.0 & 0.0 & 12.2 & 20.6 & 20.4 & 1.7 & 1.6 & 4.4 & 8.2 \\
OpenVLA & 53.7 & 0.0 & 3.0 & 22.2 & 10.6 & 19.4 & 17.6 & 28.3 & 14.3 \\
$\pi_{0.5}$ + ROCKET & 89.2 & 25.3 & 38.4 & 11.0 & 77.0 & 29.4 & 23.8 & 71.5 & 36.7 \\
\bottomrule
\end{tabular}%
}
\end{minipage}
\end{table*}
\begin{figure*}[t]
    \centering
    \includegraphics[width=\textwidth]{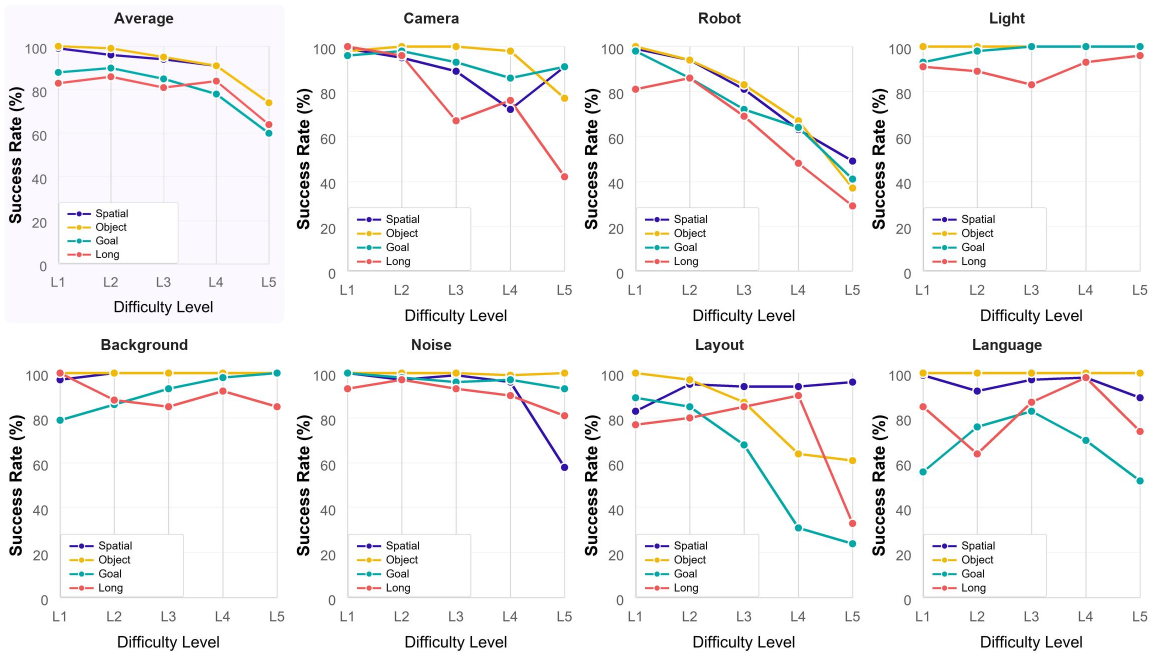}
    \caption{
    \textbf{Detailed zero-shot robustness on LIBERO-Plus.}
    We report success rates across difficulty levels L1--L5 for each perturbation category in the LIBERO-PLUS benchmark.
    The Average panel summarizes performance across all perturbation categories.
    }
    \label{fig:perturb_robustness}
\end{figure*}
\begin{table*}[t]
\centering
\caption{
\textbf{Per-task success rates on LIBERO Original.}
Task names are abbreviated by removing suite-level repeated context.
}
\label{tab:libero_original_per_task_sr_compact}
\small
\setlength{\tabcolsep}{4pt}
\renewcommand{\arraystretch}{1.02}
\resizebox{\textwidth}{!}{
\begin{tabular}{lc lc lc lc}
\toprule
\multicolumn{2}{c}{\textbf{Spatial} \; \textbf{98.6\%}} &
\multicolumn{2}{c}{\textbf{Object} \; \textbf{99.0\%}} &
\multicolumn{2}{c}{\textbf{Goal} \; \textbf{97.4\%}} &
\multicolumn{2}{c}{\textbf{Long} \; \textbf{94.6\%}} \\
\cmidrule(lr){1-2}\cmidrule(lr){3-4}\cmidrule(lr){5-6}\cmidrule(lr){7-8}
Task & SR & Task & SR & Task & SR & Task & SR \\
\midrule
Between plate \& ramekin & 100\% & Alphabet soup & 100\% & Open middle drawer & 100\% & Soup + tomato sauce to basket & 96\% \\
Next to ramekin & 100\% & Cream cheese & 98\% & Bowl on stove & 100\% & Cream cheese + butter to basket & 100\% \\
From table center & 98\% & Salad dressing & 100\% & Wine bottle on cabinet & 96\% & Stove on + moka pot on stove & 98\% \\
On cookie box & 100\% & BBQ sauce & 100\% & Top drawer open + bowl inside & 90\% & Bowl in bottom drawer + close & 100\% \\
In top drawer & 98\% & Ketchup & 98\% & Bowl on cabinet & 98\% & Mugs to left/right plates & 80\% \\
On ramekin & 96\% & Tomato sauce & 94\% & Plate to front of stove & 98\% & Book to caddy back compartment & 100\% \\
Next to cookie box & 100\% & Butter & 100\% & Cream cheese in bowl & 100\% & Mug to plate + pudding to plate & 96\% \\
On stove & 100\% & Milk & 100\% & Turn on stove & 100\% & Soup + cream cheese box to basket & 96\% \\
Next to plate & 94\% & Chocolate pudding & 100\% & Bowl on plate & 92\% & Both moka pots on stove & 88\% \\
On wooden cabinet & 100\% & Orange juice & 100\% & Wine bottle on rack & 100\% & Mug to microwave + close & 92\% \\
\bottomrule
\end{tabular}
}
\end{table*}

Table~\ref{tab:libero-plus-2x2} provides the suite-by-perturbation breakdown for LIBERO-Plus~\cite{fei2025libero}, expanding the aggregate LIBERO and LIBERO-Plus results reported in the main paper. Figure~\ref{fig:perturb_robustness} further breaks down LIBERO-Plus performance by perturbation difficulty level, showing how GAM behaves as each perturbation becomes more severe. Table~\ref{tab:libero_original_per_task_sr_compact} reports task-wise success rates on the original LIBERO~\cite{liu2023libero} benchmark.

Overall, these breakdowns show that GAM preserves strong performance on the original LIBERO tasks while improving robustness on LIBERO-Plus, especially under perturbations that require stable geometric understanding such as camera-viewpoint changes.

\subsection{Generated Future Depth Maps}
\label{sec:appendix-generated_depth}
\begin{figure*}[t]
    \centering
    \vspace{-0.35em}

    \begin{subfigure}[t]{0.95\textwidth}
        \centering
        \includegraphics[width=0.95\linewidth]{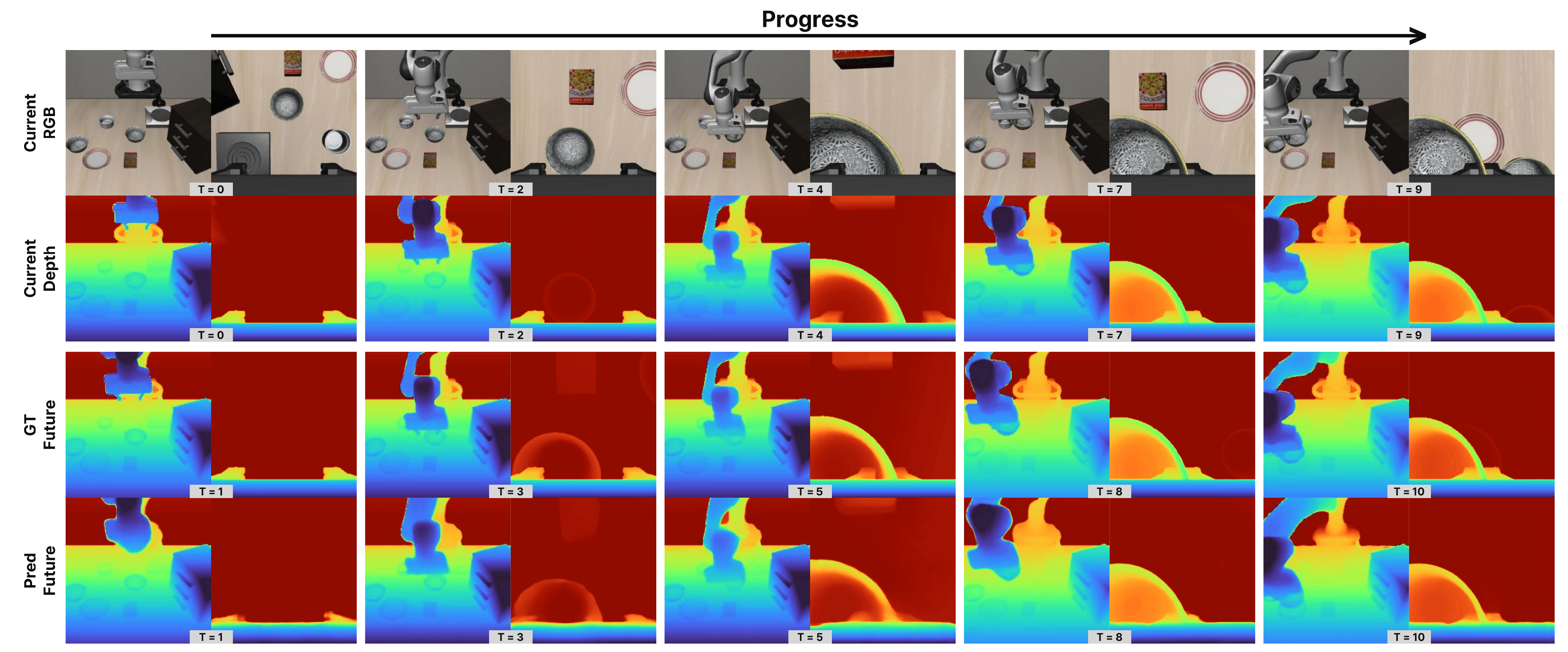}
        \caption{LIBERO-Spatial: bowl from table center to plate.}
        \label{fig:libero-spatial-depth-1}
    \end{subfigure}

    \vspace{-0.35em}

    \begin{subfigure}[t]{0.95\textwidth}
        \centering
        \includegraphics[width=0.95\linewidth]{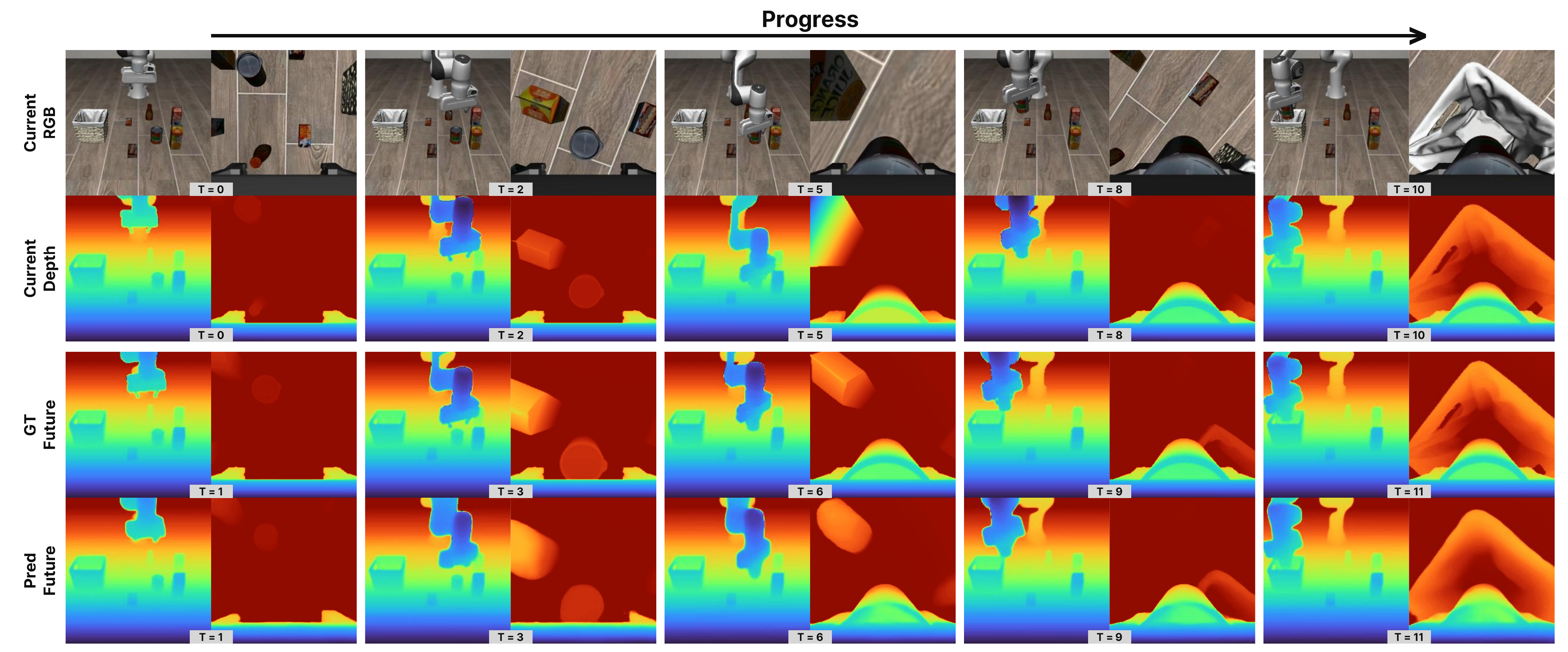}
        \caption{LIBERO-Object: tomato sauce to basket.}
        \label{fig:libero-object-depth-1}
    \end{subfigure}

    \vspace{-0.35em}

    \begin{subfigure}[t]{0.95\textwidth}
        \centering
        \includegraphics[width=0.95\linewidth]{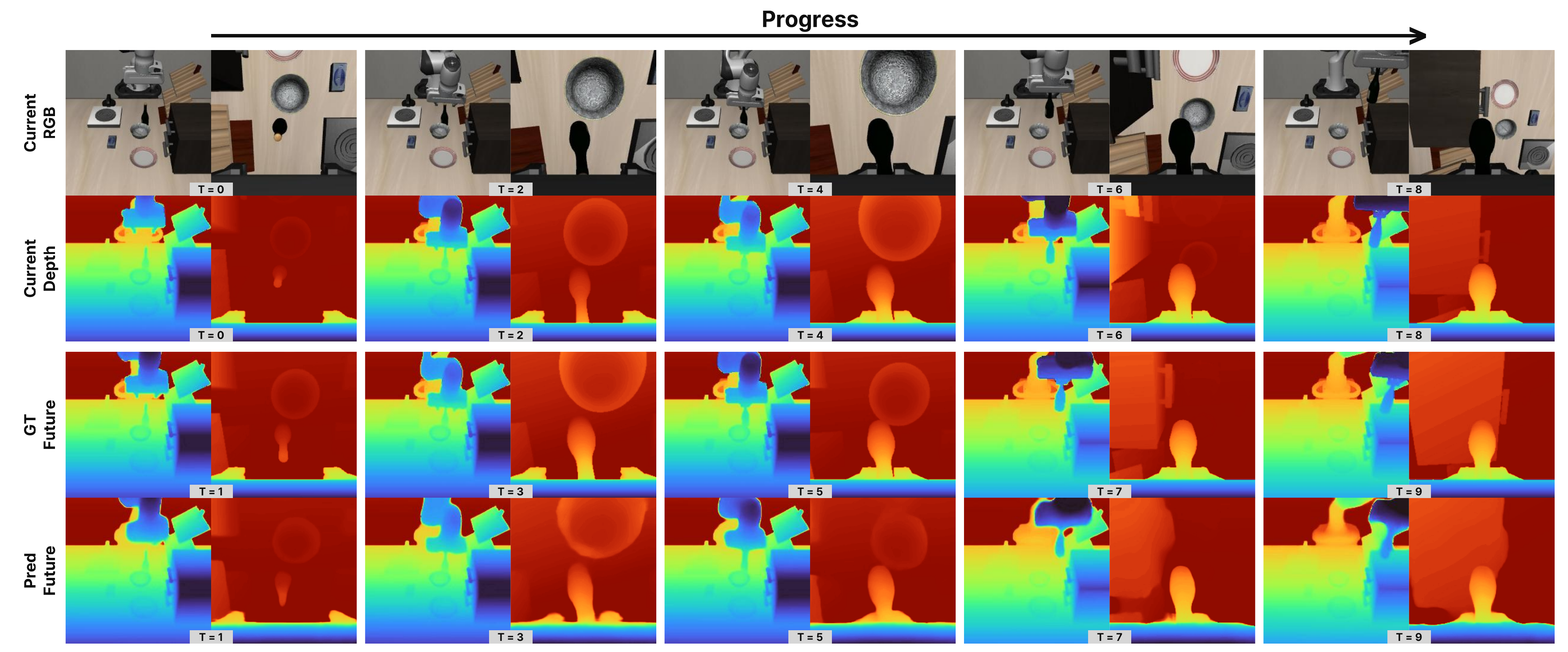}
        \caption{LIBERO-Long: cream cheese and butter to basket.}
        \label{fig:libero-long-depth-1}
    \end{subfigure}

    \vspace{-0.35em}

    \begin{subfigure}[t]{0.95\textwidth}
        \centering
        \includegraphics[width=0.95\linewidth]{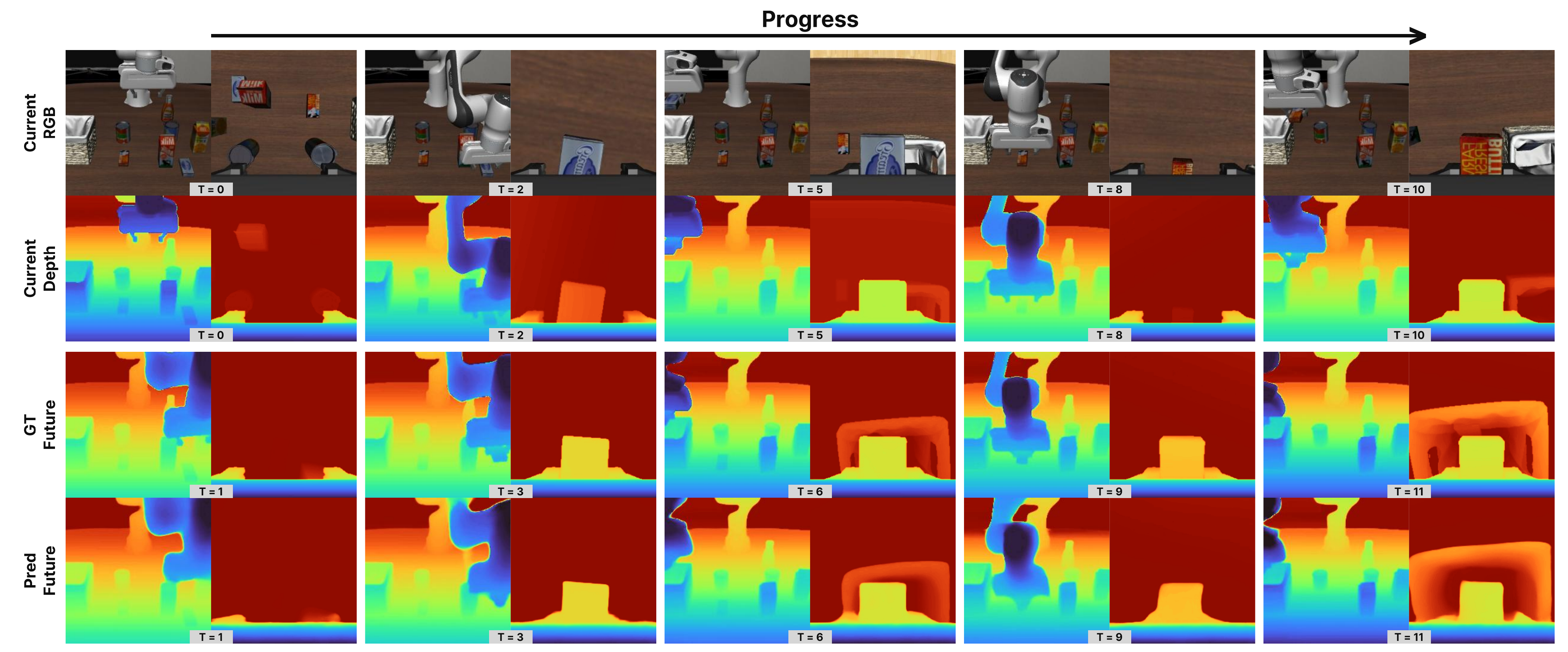}
        \caption{LIBERO-Goal: wine bottle on cabinet.}
        \label{fig:libero-goal-depth-1}
    \end{subfigure}
    \vspace{-5px}
    \caption{
    \textbf{Future depth visualizations predicted by our model on representative LIBERO tasks.}
    }
    \label{fig:libero_future_depth_visualizations}
\end{figure*}

In Figure~\ref{fig:libero_future_depth_visualizations}, we visualize the future depth predictions generated by GAM across each task suite in the LIBERO benchmark. Given a current RGB observation, GAM predicts the and future depth maps while simultaneously generating actions that align spatially with the anticipated future geometry. As demonstrated in the visualizations, GAM accurately forecasts the future depth alongside its corresponding action sequence.

\clearpage
\section{Ablation and Diagnostic Analyses}
\label{sec:appendix-ablation_analysis}

\subsection{When to Predict Actions?}
\label{sec:appendix-layer_ablation}

\begin{wraptable}{r}{0.43\textwidth}
    \vspace{-10pt}
    \centering
    \footnotesize
    \setlength{\tabcolsep}{3.5pt}
    \caption{\textbf{Direct action-token ablation.}}
    \label{tab:action_decoding_path_ablation}
    \begin{tabular}{lccc}
    \toprule
    Variant  & Orig. & Plus \\
    \midrule
    Direct-action supervision & 98.4 & 84.1 \\
    \textbf{GAM (Ours)}  & 99.6 & 89.7 \\
    \bottomrule
    \end{tabular}\vspace{-10pt}
\end{wraptable}

We additionally evaluate a direct-action supervision variant that applies the action loss directly to the output action token of the causal future predictor, without passing the action token through the remaining DA3 blocks. This ablation uses the same setting as the main component ablation on LIBERO-Object.
 
Passing the action token through the deep geometric decoder provides an additional improvement, particularly on LIBERO-Plus Object, suggesting that the remaining GFM layers contribute to refine the action representation, especially under camera perturbations.

\subsection{Attention Analysis}
\label{sec:appendix-attention_analysis}
\begin{figure*}[t]
    \centering
    \includegraphics[width=\textwidth]{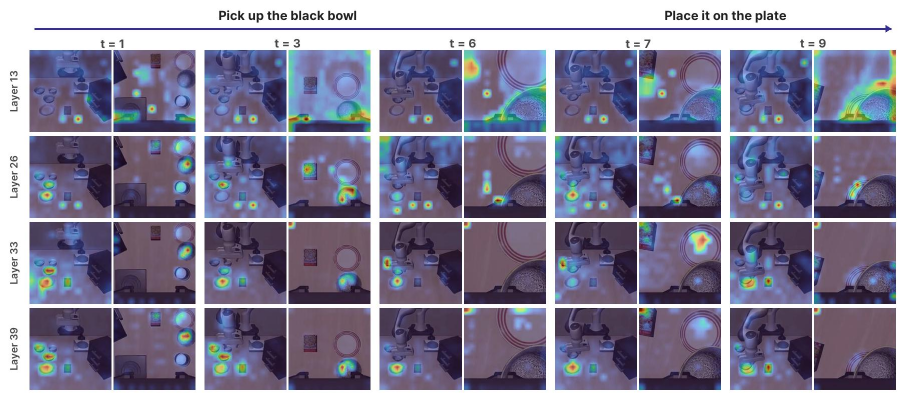}
    \caption{
    \textbf{Attention visualizations of action tokens.}
    }
    \label{supfig:attention}
\end{figure*}

In Figure~\ref{supfig:attention}, we visualize the attention maps of action tokens across GFM layers to inspect which visual regions contribute to action decoding. As shown in the attention maps, several intermediate layers attend to task-relevant regions, with clear saliency around manipulated objects and nearby contact regions. This qualitative trend is consistent with the layer ablation: mid-level representations retain object-level structure while still leaving enough depth in the GFM decoder for action-token refinement.

\clearpage
\bibliography{articles}  
\end{document}